\documentclass{ieeeaccess}

\usepackage{amsmath,amssymb,mathtools}

\usepackage{bm}
\makeatletter
\AtBeginDocument{\DeclareMathVersion{bold}
\SetSymbolFont{operators}{bold}{T1}{times}{b}{n}
\SetSymbolFont{NewLetters}{bold}{T1}{times}{b}{it}
\SetMathAlphabet{\mathrm}{bold}{T1}{times}{b}{n}
\SetMathAlphabet{\mathit}{bold}{T1}{times}{b}{it}
\SetMathAlphabet{\mathbf}{bold}{T1}{times}{b}{n}
\SetMathAlphabet{\mathtt}{bold}{OT1}{pcr}{b}{n}
\SetSymbolFont{symbols}{bold}{OMS}{cmsy}{b}{n}
\renewcommand\boldmath{\@nomath\boldmath\mathversion{bold}}}
\makeatother
\usepackage{graphicx}
\usepackage{booktabs}
\usepackage{array}
\usepackage{tabularx}
\usepackage{enumitem}
\usepackage{url}
\usepackage{cite}

\begin{document}

\history{Date of publication xxxx 00, 0000, date of current version xxxx 00, 0000.}
\doi{10.1109/ACCESS.2026.3754039}

\title{Non-Stationarity Breaks Permutation Surrogates in Multi-Agent
Reinforcement Learning: Diagnosis and Remedies}

\author{\uppercase{Nikolaos Al. Papadopoulos}\authorrefmark{1},
\uppercase{and Konstantinos E. Psannis}\authorrefmark{1}}

\address[1]{Department of Applied Informatics, University of Macedonia,
Egnatia 156, 54636 Thessaloniki, Greece (e-mail:
nikolaos.papadopoulos@uom.edu.gr; kpsannis@uom.edu.gr). ORCID: N. Al.
Papadopoulos, 0000-0003-1842-8227; K. E. Psannis, 0000-0003-0020-6394.}

\tfootnote{The publication of the article in OA mode was financially supported
by HEAL-Link.}

\markboth{Papadopoulos \headeretal: Non-Stationarity Breaks Permutation
Surrogates in Multi-Agent Reinforcement Learning}
{Papadopoulos \headeretal: Non-Stationarity Breaks Permutation Surrogates in
Multi-Agent Reinforcement Learning}

\corresp{Corresponding author: Nikolaos Al. Papadopoulos (e-mail:
nikolaos.papadopoulos@uom.edu.gr).}

\begin{abstract}
Reporting guidance for information-theoretic measures is rarely tested against ground truth.
We test one guardrail in two multi-agent reinforcement learning games, a social dilemma and a
coordination race, over 100 seeds, on agent pairs whose directed influence is zero by
construction. Omitting one precondition, the exclusion of the non-stationary training transient,
gives false-positive rates of 100.00\% and 99.95\%, flagging agents that anneal exploration
independently, in runs that never met, as influencing one another. Excluding the transient
reaches 3.05\% in the social dilemma but 11.8\% in the coordination game, so exclusion is
neither the only treatment nor a sufficient one. The remedy we recommend changes only the null
model, permuting the source within blocks of training time, and reaches 5.25\% and 5.50\%. Of
the three remedies compared (exclusion, this surrogate and conditioning on training time), only
the surrogate stays at or below the size of the test in both games. Restricted permutation is an
established construction, and the contribution lies in measuring it against ground truth. A
third null control, four learners in the social dilemma, defeats all three, and raises the rate
of conditioning by 15.20 points against 4.55 for ours. With injected links of known strength,
the surrogate detects 82.5\% where conditioning detects 58.0\% on identical pairs, while the
test without exclusion reports a positive with or without a link, so its apparent sensitivity
is uninformative. Five block counts from 16 to 256 land in the nominal region, and code and data
are released.
\end{abstract}


\titlepgskip=-21pt

\maketitle

\section{Introduction}
\label{sec:intro}

Information-theoretic (IT) measures run through modern AI practice. Shannon
entropy~\cite{shannon1948} drives information-gain splits in decision trees,
quantifies uncertainty in Bayesian neural networks~\cite{gal2016} and
regularizes maximum-entropy policies in reinforcement learning. Cross-entropy
is the default training loss for classification, and mutual information
underpins self-supervised representation learning, feature selection and the
Information Bottleneck principle~\cite{tishby2000}. Transfer
entropy~\cite{schreiber2000} is the measure this paper puts under test. It is
used to claim directed influence between components of a dynamical system,
including the agents of a multi-agent environment.

Despite this breadth of adoption, measure selection is often decoupled from
estimator assumptions, failure modes, and safe inferential claims. A
practitioner choosing between KSG and MINE for mutual information is making a
decision between a measurement estimator and a training surrogate, a
distinction with substantial consequences for how results should be
interpreted~\cite{tschannen2020}. Transfer entropy is routinely described as
measuring ``causality'' despite being a conditional MI statistic that can
arise from shared drivers~\cite{vicente2011}. KL divergence is sometimes
treated as a symmetric distance metric despite being neither symmetric nor a
true metric.

Methodological guidance aimed at these confusions normally stops at the
recommendation: the caveats are listed, and the reader is left to trust that
following them makes a measurable difference. Rather than evaluating such
guidance only conceptually, we take one specific guardrail from our own
recommendations, the requirement that transfer entropy be validated against
surrogate data only after a known non-stationary transient has been excluded,
and test it against ground truth on multi-agent reinforcement
learning systems in which the correct answer is known by construction. Where
this paper speaks of a stationarity precondition, it means that specific
operationalization: identify a transient whose source and extent are known
from the training schedule, and exclude it before pooling. We do not validate
a general stationarity-testing procedure, and nothing here should be read as
evidence about one.

The result has two parts. Omitting a single
easily-overlooked precondition, the exclusion of the non-stationary training
transient, drives the false-positive rate to $100.00\%$ and $99.95\%$ in the two
games we examined: independently trained learners that each anneal
exploration, in runs that never met, are reported as influencing one another,
in every
seed without exception. In the coordination game the ablated test also flags
$99.8\%$ of genuine pairs, so it barely discriminates at all. In this setting
the precondition therefore separates a procedure with near-nominal
false-positive control from one that effectively cannot return a negative
result.

That transfer entropy estimators assume stationarity is not a new observation.
It is well established in neuroscience, where it motivated ensemble-based
estimators that avoid pooling over time~\cite{wollstadt2014}, and the hazards
of applying these measures to collectives have been analysed
elsewhere~\cite{sattari2022}. The econometric antecedent is older: two independent series
sharing a time trend yield a significant regression statistic, which is the spurious-regression
result of Granger and Newbold~\cite{granger1974}. Three things differ in the setting studied
here. The drift is a deterministic change in the marginal. No unit root is involved, so the series
are bounded and the integration diagnostics that govern the econometric case do not describe
them. The inference is surrogate-based where the econometric case is parametric. What fails is
therefore the null the surrogate constructs. The sampling distribution of an estimator does
not enter. And the remedy that is standard
in the econometric case carries a cost specific to an information measure: differencing controls
the rate here as well as anything we recommend, but a differenced binary action series is no
longer the quantity the measure was chosen for. The remedy we arrive at is likewise not a new
object. Permuting within blocks is a restricted permutation, local to a block of training time, and
schemes of that family are already available in transfer entropy
toolkits~\cite{wollstadt2019}, as are ensemble estimators that sidestep pooling by
replicating the process~\cite{gomezherrero2015}. It is a relative of the block bootstrap of
K{\"{u}}nsch~\cite{kunsch1989}, which resamples whole blocks and preserves the order inside
each; here the blocks stay in place and the order inside them goes, which is what leaves the
profile of the drift intact. What we contribute is the
measurement: which of these remedies reaches the size of the test on a null control where
the answer is known, at what cost in sensitivity, on identical pairs.
Applying the measure to agents whose controllers change over
the analysed window is not hypothetical either: it is established practice in artificial
life, where information flow is measured in the perception-action loops of agents shaped by
evolution~\cite{klyubin2004}. Our contribution is more specific, and the mechanism
itself is not: any two independently drifting series will do this, and nothing about it
requires agents or learning. We show that directly, on synthetic
Bernoulli sequences with no agents and no game, where holding the marginal entropy exactly
fixed and removing only the trend takes the rate from $100\%$ to nominal, and lowering the
entropy further without a trend does nothing.
What multi-agent reinforcement learning supplies is a setting
where the condition holds by default and is easy to overlook. Agents that anneal
exploration carry a deterministic drift even when nothing
connects them, and direct manipulation shows that what matters is a marginal
that varies with time in both series. Synchrony between them is not required, and neither
is monotonicity, so giving each agent its own decay rate is no defence. The claim we make is
therefore about practice. The statistic itself is not in question, and the claim is
conditional: we do not
count how often transfer entropy has been applied to learning agents in the published
record, only that annealed exploration is standard, that it is sufficient on its own to
produce the failure, and that anyone who analyses agents during training therefore meets
the condition whether or not they look for it. We quantify
the consequence against a
null control where the answer is zero by construction, and show that the
remedy practitioners reach for first, changing the surrogate construction,
works only if the change is made at the right resolution.

Applying the precondition brings the rate to $3.05\%$ in the social dilemma.
That figure is indistinguishable from the floor the estimator itself returns
on stationary data at that marginal, $2.65\%$ (Fisher $p = 0.51$). The procedure adds no conservatism of its own. Placing the cutoff is not
delicate either: on the earlier eight-run pool, where the two cutoffs are paired on the same
trajectories, moving it $800$ rounds earlier into the tail of the anneal leaves that pool's
rate statistically unchanged (Section~\ref{subsec:second-game}). In the coordination game it stalls at $11.8\%$
regardless, and an equal-length window analysis shows that retreating further
from the transient does not help and in fact hurts slightly. Formal
stationarity tests go with the difference: excluding the transient leaves
$95.7\%$ of the social-dilemma series stationary but only $56.8\%$ of the
coordination series, and within the coordination game the residual
false-positive rate tracks the fraction across seeds. The four-learner social dilemma is the least stationary of the three at
$43.7\%$, so transient exclusion alone does not satisfy the stationarity
requirement there. The fraction does not order the arms, though: that same
control has the lower false-positive rate of the two measured at four
learners. We also tested the natural alternative
remedy, substituting a circular time-shift surrogate for the permutation surrogate on the
untrimmed series, and the rate did not move. Both constructions break the alignment between
series across the whole run, where the drift is steep. A third does not, and it is what we
recommend: permuting the source within blocks of training time breaks the
same alignment where the drift is flat, reaches the size of the test in both games as we
define them, and changes neither the statistic nor the data. A roster-matched third control
bounds that claim. Where the social dilemma has four learners in place of one, every remedy
we measured falls short of the size of the test, and exclusion is the lowest of the three
there, $8.20\%$ against our $9.80\%$, a difference that does not reach significance. What
survives is narrower: ours is the only one of the three at or below the size of
the test in both games, and the roster that defeats all three costs conditioning
three times what it costs ours, $15.20$ points against $4.55$. Constructing surrogates for data that are not stationary is itself an established
problem: the amplitude-adjusted and iterated schemes of Theiler et al.~\cite{theiler1992} and
Schreiber and Schmitz~\cite{schreiber2000surrogate} address the distributional side of it, and
Lucio et al.~\cite{lucio2012} build surrogates that keep a trend intact where the standard
schemes remove it. What we add to that line is a measurement, on a null control
where the correct answer is fixed by the design. Against an injected lag-1 link
it is the more sensitive of the two remedies that run on the same window, so
the construction that changes least costs no sensitivity against the one that
changes more. Against exclusion, which analyses a shorter window, the same contrast
bounds the gap instead of measuring it. Neither ordering is a property of the
procedures alone. Consider a
within-run control whose genuine influence travels on a slow ramp, where the injected link sat
at lag 1. The ramp is absorbed into the within-block null and exclusion detects more, which we
report and explain and do not average away. The construction
is not delicate about its one parameter: the five block counts we swept from 16 to 256 all
land in the nominal region. A partition derived from the stationarity test removes the
parameter altogether. Where we titrated it, against block conditioning, it matched the fixed
count on the null control at $2{,}000$ pairs and detected fewer weak links, so we report it
without recommending it. Excluding the
transient remains correct and is simpler to explain, but it discards two thirds of every run
and does not reach the size of the test in the coordination game, so it is no longer the
first thing we would tell a practitioner to do.

\subsection{Contributions}
\begin{itemize}[leftmargin=1.2em,itemsep=2pt]
  \item An experimental validation of the transfer-entropy stationarity
    guardrail against a null control, with 100 independently seeded
    replications and a burn-in ablation, replicated across two structurally
    different games (a social dilemma and a coordination race)
    (Section~\ref{sec:experiment}). This quantifies a concrete failure mode
    for practitioners applying TE to populations of independently trained
    agents, and shows the failure reproducing across two payoff structures.
    A direct manipulation of the exploration schedule identifies the annealing
    transient as a major source and excludes schedule synchrony as the
    operative factor, and an equal-length control over three $4{,}000$-round
    windows of the same runs separates the transient from sample size,
    localizing the coordination residual to something other than the schedule.
  \item Sensitivity measured by titration, which is what
    rules out the reading that a low false-positive rate is bought by a test
    that rarely fires. Injecting links of known strength into the same drifting
    streams of both games, the within-block surrogate detects $82.5\%$ and
    $89.0\%$ at a coupling where conditioning on training time detects $58.0\%$
    and $61.0\%$, on identical pairs over the same window (exact McNemar
    $p = 3 \times 10^{-9}$ and $p = 1 \times 10^{-11}$ on pairs, and
    $p = 0.002$ on each set of ten seeds, the smallest value ten seeds
    admit). On the null control the
    two fire equally often in the social dilemma, $5.25\%$ against $5.05\%$, and
    ours fires less often in the congestion game, $5.50\%$ against $8.55\%$. The
    construction that changes least is therefore also the more sensitive of the
    two, and by
    the wider margin in the game that needs a remedy
    (Section~\ref{subsec:titration}). A paired positive
    control on the real roster recovers 593 of 600 structurally connected
    pairs. We report that as corroboration, and the sensitivity evidence rests
    elsewhere, because those links run through a public signal that is a common
    input as well as the path between the endpoints
    (Section~\ref{subsec:te-result}).
  \item A result on the null model, negative for the two standard constructions and
    positive for a third. A circular time-shift surrogate leaves the false-positive rate at
    $100\%$ and $98.2\%$ (Section~\ref{subsec:surrogate}), but permuting the source
    within blocks of training time reaches $5.25\%$ and $5.50\%$, the only one of the
    four constructions we compare that is at the size of the test in both games, and it
    needs no conditioning set, no
    discarded rounds and no change of estimand (Section~\ref{subsec:conditional}).
  \item A measurement of the trend-removal baselines a practitioner would try
    first. We measured them. First
    differences control the rate as well as anything we recommend, at
    $5.0\%$, but estimate a different quantity, while re-thresholding a
    detrended binary series is the identity map
    (Section~\ref{subsec:detrend}).
  \item An end-to-end measurement of our own final recommendation, otherwise
    supported only by a correlation, together with the operating point of the
    rule it prescribes. The gate has exactly six settings, and evaluating all
    of them shows that requiring five of six series to pass admits 22 of 100
    seeds and $440$ pairs, whose rate of $6.36\%$ is not distinguishable from
    the size of the test, while requiring only four leaves the rate at $9.5\%$
    and unambiguously above it (Section~\ref{subsec:stationarity}).
  \item A comparison of $I(\mathrm{observation};\mathrm{action})$ across three
    architecturally distinct agent types (tabular RL, fixed rule, evolved
    recurrent network) under one shared protocol, showing that a
    single-run MI value for a \emph{learned} policy characterizes the run. It
    does not characterize the algorithm.
  \item The background needed to place that result, covering the four IT
    measures most often used in AI (Entropy, KL divergence and cross-entropy,
    Mutual Information, Transfer Entropy). Each is treated through the three
    questions the experiment depends on: what the measure answers, which
    estimator suits the data regime, and what its most consequential misuse
    is (Appendix~\ref{sec:background}). To make those estimator
    choices usable we supply a decision table and a
    minimum reporting checklist, which we then apply to our own experiment
    item by item (Appendix~\ref{sec:synthesis} and Section~\ref{sec:experiment}).
    These are working aids. We do not present them as validated instruments.
  \item Full reproducibility: all analysis code, generated data, and the
    seed sweep are released, and the headline mutual-information values, with
    a spot check of transfer-entropy pairs, are re-derived by an independently
    written implementation and cross-checked against a third-party toolkit.
    The conditional routine that computes the recommendation is verified
    separately, since it is our own code where the rest is the platform's: with
    a constant conditioning variable it reduces exactly to both bivariate
    implementations, and under genuine conditioning it agrees with a second
    joint encoding and returns the estimator's floor when the conditioning
    variable renders the pair independent.
\end{itemize}

\subsection{Scope}
The measure-by-measure material is background, and it is in
Appendix~\ref{sec:background} for that reason. It exists to
establish the estimator choices, assumptions and guardrails on which the experiment
depends, and Section~\ref{sec:experiment} cites it at each point where it does.
The selection table and the reporting checklist are in
Appendix~\ref{sec:synthesis} on the same principle: they are working aids the
experiment is held to. Neither is a result. Readers seeking comprehensive treatments of
individual measures are referred to the estimator-comparison studies cited
throughout.

Several related quantities are deliberately excluded. Algorithmic measures
(Kolmogorov complexity, Minimum Description Length) address model selection
and compression under a different inferential paradigm and have no
sample-based estimator in the sense used here. Directed information and
partial information decomposition remain active research areas without
consensus estimators for continuous data. System-level integration measures
(integrated information $\Phi$, effective information, autonomy) require
fully specified interventional transition probability matrices and are
restricted to small discrete systems. What puts them outside a paper about
estimating from sampled trajectories is that availability condition. The measures
themselves are not at issue, since only one of the three architectures here
exposes such a matrix, and none of the agents a practitioner would deploy at
scale does.

The citations in Appendix~\ref{sec:background} are selected on the same principle. A source
appears if it introduces or formally defines a measure or estimator,
characterizes an estimator's bias or variance empirically, or documents an
application of the kind a practitioner would recognize. We state the criterion
so that readers can see what each recommendation rests on.

\section{Experimental Validation}
\label{sec:experiment}

Appendices~\ref{sec:background} and~\ref{sec:synthesis} state recommendations.
This section tests one of them against ground truth, and reports what the
other measures return on the same system.

The results below sit at two levels of precision, and we state which is which
and do not leave it to be inferred. The failure itself, the remedies we
recommend, and the controls that bound them are measured over 100 seeds and
$2{,}000$ pairs per condition, where a rate is resolved to about one percentage
point. Four arms are smaller than that, and each says so where it appears. The single-run
control of Section~\ref{subsec:withinrun} draws 600 pairs because one run holds
no more; the non-monotone drift control of
Section~\ref{subsec:stationarity} was run at 400 pairs over 20 seeds, which is
a choice and not a limit, since the synthetic arms beside it in that section
run at the full size; the continuous-action replication of
Section~\ref{subsec:conditional} has 40 pairs; and the ensemble estimator of
Section~\ref{subsec:ensemble} rests on 30 tests. The sweeps that map the space of possible remedies run at 10 seeds and
200 pairs, because each of their cells costs a full surrogate test on every
pair; they are sized to separate $100\%$ from $5\%$. Separating $6\%$ from
$5\%$ is beyond them, and we read them only at that resolution. Each sweep below states the seed and pair
count it was run at, so which level a given rate belongs to is recoverable at
the point of use.

\subsection{Design}
\label{subsec:design}

The guardrail under test is the transfer-entropy protocol of
Appendix~\ref{sec:te} and Table~\ref{tab:master-decision}: a discrete plug-in
estimator with Miller-Madow correction, evaluated for significance against
random-permutation surrogates, on series from which a known non-stationary
training transient has been excluded. We are precise about that phrasing: what
we test is the exclusion of a transient whose source is known and whose extent
is computable from the exploration schedule. A general verification of
stationarity is a different thing, and would require its own diagnostic.

\paragraph{Questions and hypotheses} We address five questions, each with the
hypothesis it was framed to test. The hypotheses are reported as they were
written. The results have not been allowed to reshape them: of the five, one
is refuted outright and two hold only in part, and we say below which and where.

\begin{itemize}[leftmargin=1.4em,itemsep=3pt]
  \item[\textbf{Q1}] Does the stationarity precondition change the conclusions
    a practitioner would draw, on data where the correct answer is known?
    \textbf{H1}: omitting it inflates the false-positive rate above the nominal
    significance level, because a training transient leaves a trend in every
    agent's action series.
  \item[\textbf{Q2}] Is the effect a property of the training schedule or of
    the task? \textbf{H2}: the schedule, in which case the effect must
    reproduce at comparable magnitude in a game with unrelated payoff
    structure.
  \item[\textbf{Q3}] Can the failure be repaired by changing the null model
    and leaving the data alone? \textbf{H3}: yes, because a circular time-shift
    surrogate preserves each series' own temporal structure, including its
    trend, and breaks only the alignment between series.
  \item[\textbf{Q4}] Can the drift be conditioned on with nothing removed, and
    can the conditioning be parameterized without an arbitrary choice?
    \textbf{H4}: yes on both counts, because the driver is observed, with
    nothing latent about it, and because the stationarity test that diagnoses
    the problem can also serve as the stopping rule for partitioning.
  \item[\textbf{Q5}] How strong must a genuine link be before the corrected
    procedures detect it, and does excluding two thirds of the data cost
    sensitivity relative to conditioning on all of it? \textbf{H5}: exclusion
    is the less sensitive of the two at equal calibration, because it answers
    the same question from a third of the sample.
\end{itemize}

H1 and H2 were supported. H5 holds over part of the range only: exclusion is
indeed the less sensitive at matched calibration in the mid range, and the
ordering does not survive at the weakest couplings
(Section~\ref{subsec:titration}). H3 was refuted and H4 is partial, and both
need their answers separated from their hypotheses.

Q3 asks whether
the failure can be repaired by changing the null model, and the answer is yes:
that is the construction this paper ends up recommending. H3 named the wrong
change. A circular time shift does not help
(Section~\ref{subsec:surrogate}), because it breaks the correspondence between
the series across the whole run where the drift is steep; permuting within
blocks breaks it where the drift is flat, and works
(Section~\ref{subsec:conditional}). The hypothesis was refuted and the question
it was asked about was answered affirmatively, by a mechanism we did not
anticipate when we wrote it.

H4 was supported in one configuration of two and we count it as partial.
Conditioning on training time does bring the untrimmed rate to the size of the
test in the one-learner social dilemma, and the partition can indeed be
derived from the stationarity test. No value has to be chosen by hand, which is the second half of H4. But
the same
conditioning reaches $8.55\%$ in the coordination game and stops there
(Section~\ref{subsec:conditional}), so the first half holds where the schedule
is what the series are drifting with. Elsewhere it does not.

\paragraph{System} The primary experiment uses the repeated $N$-player Public
Goods Game on the open-source GameBrains multi-agent
platform~\cite{gamebrains2026}, used here strictly as a library. Three
architecturally distinct agent types share the protocol: two
independently-seeded tabular Q-learning agents; three fixed strategies
(always-cooperate, always-defect, majority tit-for-tat); and, in a separate
arm, a population of Markov-brain animats evolved by a generational genetic
algorithm~\cite{albantakis2015}. Each agent observes only a public signal, the
number of cooperators in the previous round. No agent observes another agent's
individual action.

That observation structure makes per-pair ground truth available in
this game. Any apparent directed influence between two Q-learners must run
through the shared public signal, never through direct observation, and
pairs involving a constant strategy cannot carry information at all. Pairs
assembled from separate runs, the null control below, cannot be linked by any
route. The null control is the stronger of the two, because its ground truth
does not depend on reasoning about the observation channel: two agents that
never shared a game have zero directed influence by construction.

\paragraph{Conditions} The design is a $2 \times 2$: two data conditions
crossed with two analysis procedures.
\begin{itemize}[leftmargin=1.2em,itemsep=2pt]
  \item \emph{Real roster}: the five interacting agents above, where genuine
    signal-mediated dependencies exist.
  \item \emph{Known-null control}: 20 ordered agent pairs per seed assembled
    post hoc from independently seeded, non-interacting runs. True directed
    influence is zero by construction, so every significant result here is a
    false positive. Every calibration figure in this paper draws on one pool:
    six independently seeded runs per replication, 20 ordered pairs from them,
    100 replications, in both games. An earlier pool of eight runs and 30
    ordered pairs indexed its runs so that they collide across replications,
    and Section~\ref{subsec:second-game} states what still rests on it and
    what does not. Each table names the pool it used. In every case the pairs
    are drawn the same way: all ordered pairs of distinct runs are enumerated,
    permuted by a generator seeded from the replication index, and the first
    $N$ taken. The selection is therefore fixed before any statistic is
    computed and reproduces exactly on a re-run.
  \item \emph{Transient-exclusion procedure}: analysis restricted to the
    post-transient tail (rounds 8{,}000 to 12{,}000). The cutoff is derived
    from the schedule: $\epsilon$ reaches its floor of $0.02$ at round
    $7{,}823$, so exploration is constant over the analysed window. Constant
    exploration removes one known source of non-stationarity. It does not
    establish that the process is stationary.
  \item \emph{Ablation}: the identical pipeline on the identical runs with
    the burn-in exclusion switched off, i.e.\ the full 12{,}000-round series.
\end{itemize}

Because the two controls are labelled in opposite directions while both
involve a common driver, it is worth stating precisely what is being called
correct in each. The surrogate test asks whether the source series is
exchangeable with respect to the target's future. On the null control the two
agents occupy separate runs, so no path of any kind connects them and the
structural answer is zero; the schedule is a driver of both, but it is an
artefact of the experiment. The system contains no such channel. On the real
roster the public signal is a mediator. It is not a common cause, because an agent's action
enters the cooperator count, which the others observe on the next round, so a
path exists and information genuinely traverses it. Conditioning on a mediator
would remove that dependence, which is why doing so would not show it to be
spurious. What the ablation's $100\%$ therefore measures is not a
miscalibrated test. Two series pooled across a shared trend are genuinely
non-exchangeable, and the test is right to say so. It is the rate at which a
statistically correct answer fails to answer the structural question that was
asked. What fixes it is therefore a treatment of the non-stationarity itself,
applied either to the data or to the null the surrogates build. A test with
better small-sample behaviour would change nothing.
Every cell of this design is replicated over the same 100 seeds, and every
calibration claim in this paper rests on that replication level. The parameter
sweeps reported later use 10 or 20 seeds, giving 200 to 400 pairs per cell.
That is a deliberate choice. It is not an oversight. Their purpose is to locate a rate
by order of magnitude, and 200 pairs place a rate of $6\%$ in a $95\%$ Wilson
interval of $[3.5\%, 10.2\%]$, far from the $[98.1\%, 100\%]$ of an
unconditional cell. The same 200 pairs cannot separate $6\%$ from the nominal
$5\%$, and we do not use them for that. Where the distance to nominal is the
claim, as in the main $2 \times 2$, $2{,}000$ pairs narrow the interval
to $[2.38\%, 3.90\%]$. Each table states its own seed count and pool.
Significance uses 200
random-permutation surrogates per directed pair at $\alpha = 0.05$: the
source's action series is shuffled uniformly at random while the target's is
left intact, and the observed transfer entropy is compared against the
resulting null distribution. This is the most widely used surrogate in the
discrete case, and it is the specific construction whose blind spot the
ablation exposes.

\paragraph{Statistical analysis} Two levels of inference are involved and it
is worth separating them. The surrogate $p$-value decides whether an
individual pair is flagged. The analysis below concerns how often that
decision is wrong across pairs.

The design contains two dependencies that dictate which tests are admissible.
First, the two procedures are applied to the identical control pairs within
each seed, differing only in which slice of the series is used, so that
comparison is paired and we test it with an exact McNemar test on the
discordant pairs. Second, and for the same reason, the three equal-length
windows are repeated binary measurements on the same pairs, so we use
Cochran's Q. A $\chi^2$ test on pooled counts would treat the three
measurements as independent samples.

A third dependency cannot be removed by choice of test and we therefore
report around it. Control pairs are assembled from a smaller pool of
trajectories, so a given trajectory appears in several pairs and the
pair-level outcomes are clustered, and are not independent Bernoulli trials.
Pair-level rates and their Wilson intervals are accordingly reported as
descriptive summaries, and every inferential claim about a rate is
accompanied by a seed-level analysis, in which the seed is unambiguously the
independent unit: one rate per seed, summarized by a mean and a small-sample
$t$-based interval over the 100 seed-level values. With 100 independent
seeds this interval is narrow enough to be read as an inferential statement
in its own right, and we use it to check whether each pair-level conclusion
survives at the level where independence is guaranteed. Where the two levels
disagree we say so. Wilson intervals are used at the pair level because they
remain defined at the boundary, where the normal approximation collapses to
zero width. The paired tests, exact McNemar and Cochran's $Q$, are computed at
the pair level without the cluster correction the same argument would demand,
which we accept because the contrasts they address span an order of magnitude
and no plausible design effect changes them. That exemption is granted on a
premise, so we checked it. Every contrast the paper
draws below an order of magnitude was recomputed with the seed as the unit,
from the stored per-pair records: conditioning over exclusion at $c = 0.04$
moves from $p = 0.034$ at the pair level to $p = 0.023$ by Wilcoxon signed rank
over the ten seed-level rates, the within-block surrogate over conditioning at
$c = 0.04$ from $3 \times 10^{-9}$ to $0.002$, and over exclusion at $c = 0.02$
from $0.008$ to $0.008$. The game difference at matched roster moves from
Fisher $p = 2 \times 10^{-4}$ to Mann-Whitney $p = 0.003$ over the 100
seed-level rates, and the derived partition against the fixed count at
$c = 0.02$ from exact McNemar $p = 0.007$ to Wilcoxon $p = 0.008$ over the
ten. None reverses, and we report the seed-level values
where they carry a claim. Exact binomial tests assess consistency with the
size the test actually realizes with a finite surrogate count, $10/201 =
4.98\%$, and every one of them in this paper is computed against that rather
than against the nominal $0.05$. The two differ enough to matter only where a
rate sits near the threshold, and Section~\ref{subsec:te-result} works through
the one rate that sits there. The verdict is the same under either target; what
changes is only which figure the comparison is honest about. All tests are
computed by a released script from the recorded per-pair outcomes, with two
exceptions named where they fall: the weighted-block arm of
Section~\ref{subsec:conditional} did not retain them, so its comparison is
unpaired and we read it as the weaker test it is, and the cutoff contrast of
Section~\ref{subsec:second-game} retained per-seed counts and its discordant
totals rather than the outcomes themselves, so its McNemar figure is read from
the stored summary.

\subsection{Transfer Entropy: The Guardrail Under Test}
\label{subsec:te-result}

\begin{figure}[htbp]
\centering
\includegraphics[width=\linewidth]{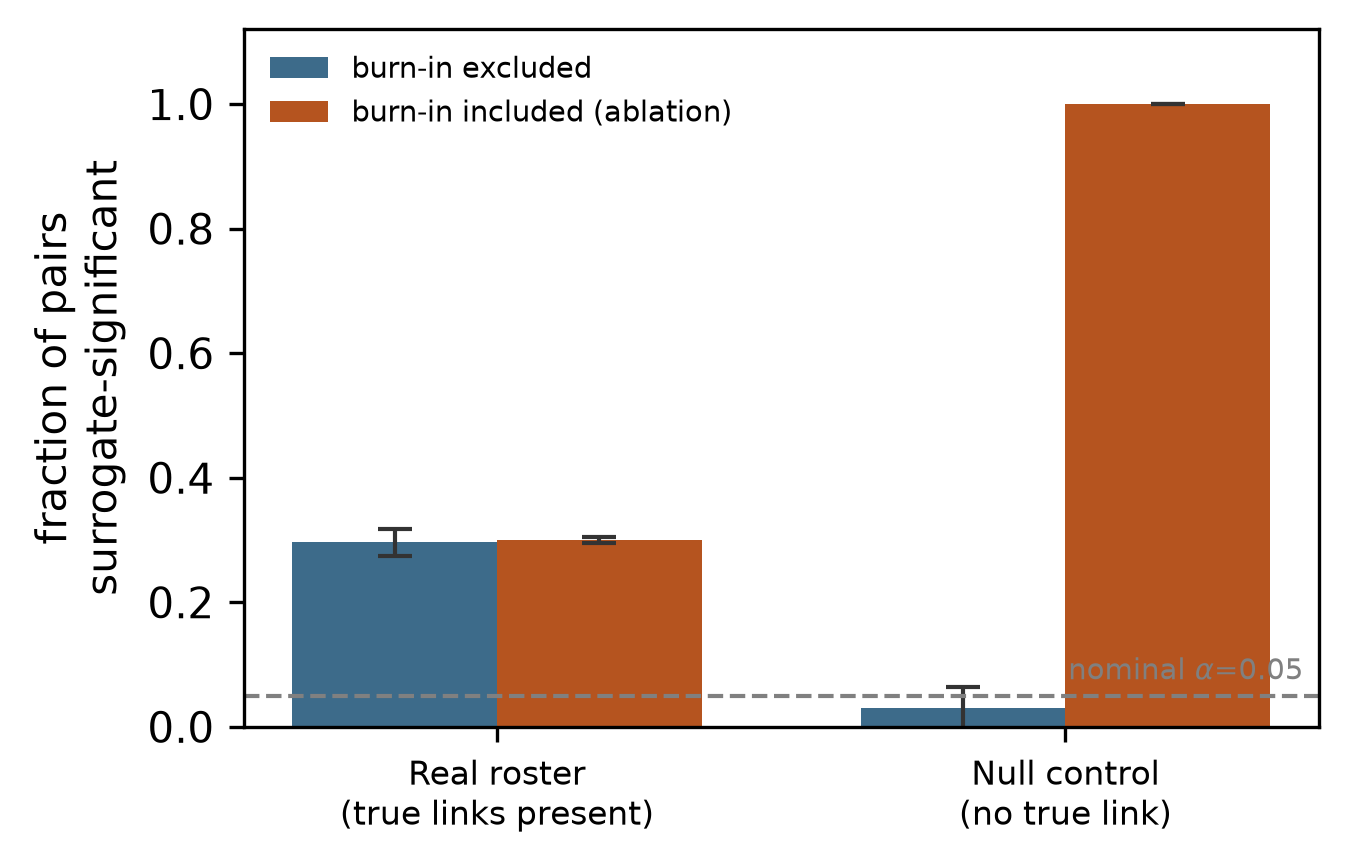}
\caption{Public Goods Game: fraction of directed agent pairs flagged as
  surrogate-significant, mean $\pm$ standard deviation over 100 seeds. On the
  real roster the two procedures are near-indistinguishable, recovering the
  same six structural links. On the null control, where no true link exists by
  construction, transient exclusion sits at $3.05\%$, which is indistinguishable
  from the floor the estimator returns on stationary data at this marginal,
  while the ablation
  flags every one of the $2{,}000$ pairs
  ($100\%$). Excluding the non-stationary transient therefore costs almost none
  of the sensitivity to these six structural links and removes essentially all
  false positives. What it costs against weaker links of known strength is
  measured in Section~\ref{subsec:titration} and is not small. Table~\ref{tab:two-games} reports
  the same comparison for the congestion game. The real-roster bars are one match per
  replication; the null-control bars are the six-run pool of
  Section~\ref{subsec:design}, 20 ordered pairs per replication over 100 replications.
  \texttt{experiments/make\_figure1.py} redraws this figure and prints the stored file and
  field behind each of the four bars.}
\label{fig:te-validation}
\end{figure}

Figure~\ref{fig:te-validation} reports the result. Three observations follow.

\emph{The procedure recovers the true structure: a positive control.} A test
that never fires would score perfectly on a null control, so calibration
evidence is worthless without evidence of sensitivity. The real roster
supplies it, because its per-pair ground truth is known from the observation
structure. Six of the 20 directed pairs are structurally connected through the
public signal. Transient exclusion recovers $593$ of the $600$ connected
pairs across the 100 seeds, $98.8\%$.

The other 14 pairs have a constant strategy at one endpoint, and we report no
specificity figure from them because none is obtainable. Permuting a constant
series returns that same series, so every surrogate reproduces the observed
statistic exactly; the surrogate distribution collapses to a point mass, with
standard deviation $0$ in all $1{,}400$ cases, and the $p$-value is $1.0$
whatever the transfer entropy happens to be. The statistic is degenerate as
well as the test: a constant variable has zero entropy, so
$I(X_{t+1}; Y_t \mid X_t) \leq \min\{H(X_{t+1}), H(Y_t)\} = 0$ exactly, and the
measured values bear that out at $0$ and $2 \times 10^{-16}$~bits for the two
endpoints. Those pairs cannot be flagged
under any estimator, and counting them as correct rejections would measure the
construction. It would say nothing about the procedure. The calibration
evidence in this paper
rests on the null control, where the pairs carry genuine variation and could in
principle be flagged.

Sensitivity of $98.8\%$ is likewise a property of this test set. It is not a
demonstrated property of the procedure, and for two reasons. Every genuine
link here is of similar strength, which Section~\ref{subsec:titration} addresses by
varying that strength deliberately. And the six links run through the public signal, which
is a common input to both endpoints as well as the path between them, so a test that fires
because two agents received the same observation scores as a hit on this roster. Given that
the paper's thesis is that shared inputs produce detections answering the wrong structural
question, this control cannot separate sensitivity to influence from sensitivity to shared
input. The mediator argument of the design establishes that a path exists. Whether the
path is what fired is a separate question. The titration is the clean answer to both objections, since it injects
links into streams that share nothing at all. What the roster does license is reading the
null-control result as calibration when it may be inertness: a
test that recovers almost every true link cannot be dismissed as one that
simply never fires. The seven misses are concentrated in two of the six links,
which are also the two weakest.

The six recovered links are exactly the structurally connected ones: both directions between the two
Q-learning agents ($T_{Q_0 \to Q_1} = 0.105$~bits, $T_{Q_1 \to Q_0} =
0.095$~bits at seed 0), and both directions between each Q-learner and the
majority-tit-for-tat agent, which by construction is a deterministic function
of the public signal ($T_{Q_0 \to \mathrm{TFT}} = 0.537$,
$T_{Q_1 \to \mathrm{TFT}} = 0.411$, $T_{\mathrm{TFT} \to Q_0} = 0.121$,
$T_{\mathrm{TFT} \to Q_1} = 0.038$~bits; all $p = 0.005$). The remaining 14
pairs involve a constant strategy at one endpoint and measure exactly zero
bits ($p = 1.0$): a deterministic constant action carries no information to
detect.

\emph{Applied as the guardrail requires, transient exclusion is calibrated on
this null control.} It flags 61 of
$2{,}000$ pairs pooled across seeds, $3.05\%$, Wilson $95\%$ CI
$[0.0238, 0.0390]$. Comparing that to a nominal $0.05$ would be comparing
against the wrong target, and the correction is worth making even though here
it does not change the verdict. With $S$ surrogates the attainable $p$-values
are multiples of $1/(S+1)$, so a threshold of $0.05$ actually rejects when at
most $9$ of the $200$ surrogates match or exceed the observed statistic, and
the realized size of the test is $10/201 = 0.0498$. The nominal $0.05$ is not
attainable. Against
that figure an exact binomial test gives $p = 3 \times 10^{-5}$. It would be
easy to read that as conservatism in the procedure, and we report the check
that says otherwise.

The post-transient marginal is about $p = 0.010$, so a $4{,}000$-round window
carries roughly forty ones across an eight-cell joint table. Ties in the
surrogate distribution and near-empty cells both push a permutation test
towards under-rejection, and neither has anything to do with excluding a
transient. To separate the two we generated, for every seed, six stationary
Bernoulli series whose marginals match that seed's six real post-transient
series exactly, series for series. These are independent and stationary by
construction, so whatever they return is the floor the estimator and the
surrogate procedure impose on their own. The floor is $53$ of $2{,}000$,
$2.65\%$, CI $[2.03\%, 3.45\%]$, and the observed $3.05\%$ is not
distinguishable from it (Fisher $p = 0.51$). The shortfall below the realized
size therefore belongs to the estimator at these marginals. The guardrail does
not produce it, and we claim only that the procedure is calibrated at or below
the size of the test. We do not claim that it is conservative. Where a rate sits near
the threshold the distinction between
$0.05$ and $0.0498$ can decide the verdict, and we use the realized size
throughout.

\emph{Removing the precondition destroys that calibration.} The
ablation flags $2{,}000$ of $2{,}000$ null-control pairs, $100\%$, CI
$[0.9981, 1.0000]$, in every seed without exception. Of the $2{,}000$ paired
comparisons, $1{,}939$ are discordant in the direction where the ablation
flags and transient exclusion does not. Not one is discordant the other
way.
The exact McNemar $p$-value underflows double precision; the effect is the
$96.95$ percentage-point difference itself, with every discordant pair falling
on the same side.

A reader who disputes the framing above, and holds that a test reporting dependence between
two genuinely non-exchangeable series is not making an error, can reach the same conclusion
without conceding anything. Section~\ref{subsec:titration} injects links of controlled
strength and finds the ablated procedure reporting $100\%$ at $c = 0$, where no link
exists, and $100\%$ at $c = 0.32$, where a strong one does. A procedure whose output does
not vary with the presence of the effect it claims to detect is uninformative whatever one
makes of its individual rejections, and that argument needs no null control, no ground
truth and no reasoning about observation channels. It is the form of the result we would
ask a sceptical reader to weigh.

The mechanism is specific, and it
is invisible in the data itself. Every agent here anneals its exploration rate,
so every action series carries a deterministic downward trend. A
random-permutation surrogate destroys the temporal structure of the source
series but leaves that trend intact in the target, so the unpermuted pair is
reliably more ``predictable'' than its surrogates even when the two agents
come from runs that never interacted. Section~\ref{subsec:schedule} shows by
direct manipulation that the trend is what does the damage. Synchrony between
the schedules contributes nothing. Two agents trained independently, with no shared
environment whatsoever, are thereby incorrectly flagged as exhibiting directed
influence.

\emph{Would a magnitude cutoff have done the same job?} A reader may reasonably
ask why the false positives cannot simply be discarded for being small, which
would need neither exclusion nor conditioning. The question is answerable only
with the magnitudes, so we report them. Over 20 seeds and 400 null-control
pairs the ablation's false positives have a median transfer entropy of
$0.0119$~bits and a maximum of $0.0179$; the same pairs after exclusion
measure $0.00014$~bits, so the retained transient inflates the statistic by
roughly two orders of magnitude. Those false positives are not small. They
overlap the genuine links, whose fifth percentile is $0.0075$~bits, and the
most permissive cutoff that removes every one of them, $0.0179$~bits, discards
$21.7\%$ of the structurally connected pairs. Against that, transient
exclusion loses $7$ of $600$. A magnitude cutoff is therefore not a cheaper
substitute for the precondition on this data; it is a worse trade, and the
reason is that the drift inflates true and false values alike.

The practical consequence is asymmetric in a useful way. In this game the real
roster gives near-identical results under both procedures, $29.6\%$ of pairs
flagged under transient exclusion against $29.9\%$ under the ablation,
so applying the precondition sacrifices almost no sensitivity while
omitting it converts a calibrated test into one that cannot produce a negative
result.

That sentence combines two arms which differ in the property that most governs
detectability, and the join should be made explicit. A reader should not have
to find it. Specificity is measured on null-control streams from agents trained in isolation,
whose late-window action entropy is $0.083$~bits; sensitivity is measured on the real
roster, whose converged tail sits at $0.866 \pm 0.239$~bits, precisely because the
co-adaptation we are studying prevents the collapse. A near-deterministic series is easy to
leave unflagged and a high-entropy one is easy to flag, so neither arm on its own licenses
the composite claim. What does license it is
Section~\ref{subsec:titration}, which injects links of controlled strength into the
null-control streams themselves and so measures both properties on one population. The
figures above should be read as the two endpoints that motivated that experiment. They
are not a sensitivity-specificity pair.

\paragraph{Ruling out sample size} The comparison so far confounds two things.
The ablation analyses $12{,}000$ rounds and transient exclusion
$4{,}000$, and a permutation test gains power as $n$ grows, so more data alone
would push more pairs past the significance threshold. That is a genuine
alternative explanation for the $100\%$ rate and it cannot be argued away, so
we measured it instead.

We re-ran the null control on three windows of the \emph{same} length,
$4{,}000$ rounds each, taken from the same runs, in both games:
\emph{early} (rounds 0 to $4{,}000$, $\epsilon$ falling from $1.00$ to
$0.14$), \emph{middle} (rounds $4{,}000$ to $8{,}000$, $\epsilon$ falling from
$0.14$ to its floor), and \emph{late} (rounds $8{,}000$ to $12{,}000$,
$\epsilon$ at the floor throughout). Sample size is identical across all six
cells, so it cannot account for any difference between them.

One thing besides the transient does vary, and it belongs in the open. As
exploration anneals the policies concentrate, so
the marginal entropy of the action series falls with the window: over 20 seeds
it runs $0.752$, $0.190$, $0.083$~bits in the social dilemma. Both the plug-in
bias and the power of the permutation test depend on that marginal, so this
design separates sample size from everything else but does not separate
residual transient from concentration of the marginal.

The two games separate them for us, because the concentration does not proceed
alike. The congestion marginals run $0.766$, $0.222$, $0.185$~bits, so its late
window retains twice the entropy of the social dilemma's and, more tellingly,
fifteen times the spread across seeds ($\pm 0.134$ against $\pm 0.009$). Where
the marginal concentrates uniformly the rate falls to floor; where it stays
high and seed-dependent the rate does not. That is the same split the
stationarity diagnostics of Section~\ref{subsec:stationarity} report from the
other direction, and it is a further reason to read the congestion residual as
policies that have not settled. The test is not failing.
Table~\ref{tab:windows} reports the result.

\begin{table}[htbp]
\centering
\caption{Null-control false-positive rate over equal-length windows
  ($4{,}000$ rounds each; 20 pairs per seed, 100 seeds). Sample size is
  constant throughout, so it cannot account for the variation. Upper line per
  cell: rate pooled over the 2{,}000 pairs. Lower line: $95\%$ interval on the
  seed-level mean over the 100 seeds, the unit at which independence is
  guaranteed. Cochran's Q across the three windows gives $p < 10^{-300}$ in
  both games. The two games separate at the last step: the public-goods rate
  keeps falling, while the congestion rate rises again.}
\label{tab:windows}
\footnotesize
\setlength{\tabcolsep}{4pt}
\begin{tabular}{@{}lccc@{}}
\toprule
\textbf{Window} & \textbf{Exploration rate} & \textbf{Public Goods} &
\textbf{Congestion} \\
\midrule
early  & $1.00 \to 0.14$ & $99.5\%$ & $99.0\%$ \\
       &                 & {\scriptsize seed [99.2, 99.8]} & {\scriptsize seed [98.6, 99.4]} \\[3pt]
middle & $0.14 \to 0.02$ & $6.6\%$  & $8.1\%$ \\
       &                 & {\scriptsize seed [5.4, 7.8]} & {\scriptsize seed [6.7, 9.5]} \\[3pt]
late   & $0.02$ (floor)  & $3.05\%$ & $11.8\%$ \\
       &                 & {\scriptsize seed [2.4, 3.7]} & {\scriptsize seed [10.1, 13.5]} \\
\bottomrule
\end{tabular}
\end{table}

Sample size does not explain the effect: it is held constant here and the rate
still spans more than an order of magnitude in each game. Because the three
windows are measured on the same 2{,}000 pairs, a $\chi^2$ on pooled counts
would treat them as independent samples. We therefore use Cochran's
Q: $Q = 3{,}614$ for the Public Goods
Game and $Q = 3{,}236$ for the congestion game, $\mathrm{df} = 2$,
$p < 10^{-300}$ in both. The seed-level analysis agrees and is reported in
Table~\ref{tab:windows}.

The two games then diverge, and with 100 seeds the divergence is unambiguous
at both levels of analysis. In the Public Goods Game the relationship is
graded and reaches bottom: each step away from the transient lowers the rate,
from $99.5\%$ to $6.6\%$ to $3.05\%$, and the furthest window sits below the
realized size of the test (seed-level $95\%$ CI $[0.024, 0.037]$; exact
binomial $p = 3 \times 10^{-5}$), which
Section~\ref{subsec:te-result} attributes to the estimator at this marginal.
The window is not the cause. The congestion game behaves differently.
Its first step is just as large, $99.0\%$ to $8.1\%$, but
the second step moves the wrong way: the late window is \emph{worse} than the
middle one, $11.8\%$ against $8.1\%$, with non-overlapping seed-level
intervals. Moving further from the exploration transient does not merely stop
helping in this game. It makes matters slightly worse.

That residual is unambiguously above nominal. The seed-level $95\%$ CI is
$[0.101, 0.135]$, which excludes the size of the test, and the pair-level
exact binomial test agrees ($p = 2 \times 10^{-33}$ on 236 of $2{,}000$). The seed-level and pair-level analyses
concur.
What the window comparison establishes is that the residual is not sustained
by the exploration transient, since retreating from that transient does not
reduce it. Section~\ref{subsec:stationarity} measures how much of it
residual non-stationarity accounts for, which is part of it and not
demonstrably all.
Continued co-adaptation between agents still negotiating who yields is a
hypothesis. We do not report it as a finding, and testing it would require a
direct stationarity diagnostic or frozen-policy trajectories.

\subsection{Does the Failure Generalize? A Second Game}
\label{subsec:second-game}

The diagnosis above attributes the failure to the training schedule. The
payoff structure is not what produces it. That is a testable prediction: if it is correct,
the ablation must produce a comparable inflation in a game with entirely
different incentives. It says nothing about how completely the remedy will
work there, and as it turns out the two games differ on exactly that point.

We therefore repeated the comparison on the congestion, or Honey-Jar, game
from the same platform: an episodic coordination race in which agents advance
along a corridor and payoff depends on how many arrive together. It is a
coordination problem where the first was a social dilemma. Its state encodes a
board position where the first counted cooperators, and its episodes terminate
on arrival where the first ran as one continuous sequence. What it shares with
the first experiment is only what the diagnosis says should matter: binary
actions and an $\epsilon$-greedy schedule. Four Q-learners, 12{,}000 rounds,
100 seeds, 20 null-control pairs per seed, otherwise identical protocol.

\emph{The prediction holds, and the ablation is if anything worse.} Without
the burn-in exclusion, 1{,}998 of this arm's 2{,}000 null-control pairs are
flagged as significant, $99.9\%$, against the $99.95\%$ the same ablation
returns in the six-run pool of Table~\ref{tab:decomposition}, essentially as in
the Public Goods Game. Here the
collapse is more complete: the ablation also flags $99.8\%$ of real-roster
pairs, so the test barely separates the two conditions at all. Almost every
comparison it is given returns ``significant,'' indicating that the test has
lost discriminatory value under the ablated protocol.

\emph{The remedy reduces the inflation here too, but does not restore
nominal calibration.} Applying the precondition reduces the null-control rate
from $99.9\%$ (CI $[0.9964, 0.9997]$) to $11.0\%$ (220 of 2{,}000; CI
$[0.0970, 0.1245]$), with 1{,}778 discordant pairs all in the same direction,
exact McNemar $p < 10^{-300}$. Unlike the Public Goods Game, that rate remains
well above the nominal level (binomial $p = 4 \times 10^{-27}$).

This quantity appears three times at $2{,}000$ pairs and the three figures
differ, so we say exactly what they share. It appears twice more in the sweeps,
at $200$ pairs, where the conditioning arm's post-transient baseline reads
$11.5\%$ and the titration's null row $11.0\%$; those carry the resolution of a
sweep and are read as such. The second-game arm just described gives
$11.00\%$, the equal-length late window gives $11.80\%$, which
Table~\ref{tab:two-games} and Table~\ref{tab:windows} report because it is
measured on the same trajectories and the same pairs as the ablation beside it,
and the surrogate comparison gives $11.95\%$, CI $[0.1060, 0.1345]$
(Table~\ref{tab:surrogate}).

These are not three independent estimates and it would be wrong to read their
agreement as replication. All three build their six congestion runs from the
same generator expression at the same 100 seeds, so the $600$ trajectories are
the same trajectories, and all three analyse the same post-transient window. Two
of the three also select the same 20 ordered pairs per seed, drawing them from a
generator seeded at $s + 4242$; the second-game arm uses $s + 9999$ and so
selects different pairs from the same runs. What varies across the three is
therefore the pair selection in one case and the surrogate draw in all of them.
The data is the same throughout. The right reading is that the residual is $11$
to $12\%$ on this
set of runs, that the analysis choices which differ between the three do not
move it, and that a value measured on fresh trajectories would be a stronger
claim than any of them. The powered figure to quote is the $11.80\%$ of
Table~\ref{tab:two-games}, which shares its pairs with the ablation it is
contrasted against.

The same quantity in the Public Goods Game appears three times, collected here
so the figures are not mistaken for disagreement: $61/2{,}000 = 3.05\%$
over 100 seeds in Table~\ref{tab:two-games} and the same figure in
Table~\ref{tab:windows}, then $5.5\%$ over 10 seeds
(Section~\ref{subsec:detrend}) and $4.5\%$ over 10 through the coupling
construction (Table~\ref{tab:titration}). The same caution applies as in the
congestion game. The ten-seed figures are computed on the first ten of the same
100 seeds and therefore on the same trajectories and the same pairs, so they are
not independent estimates; and they differ from the 100-seed figure in more than
seed count, since the main arm uses the platform's transfer-entropy routine
while the two sweeps use the implementation released with this paper. The
spread across the three shows what a smaller sample and a second implementation of
the same estimand produce, and the $3.05\%$ is the figure to quote.

The single pool of Section~\ref{subsec:design} is indexed by run seeds
$100s + 10k$ for replication $s$ and run $k$, which collides for no pair of
indices, so the 600 trajectories are distinct and the 100 seed-level rates are
100 independent replications. An
earlier eight-run pool indexed its runs as $s + 100 + 10k$, which collides
whenever $s$ rises by ten as $k$ falls by one, so its 800 cells drew on 170
distinct trajectories and its replications were not independent. No calibration
figure rests on it. One later arm still draws from it, the cutoff sensitivity
of Section~\ref{subsec:second-game}, and it is a paired contrast in which both
arms carry the same structure. Where a rate from it is printed, in that
subsection and in the one-sentence summaries in the introduction and the
conclusion, it is one side of that contrast and carries the pool's dependence
inside it, so no calibration claim rests on its absolute value. Two arms that
an earlier version of this paragraph listed beside it do not draw on the pool:
the block-placement comparison of Section~\ref{subsec:conditional} builds six
runs on the non-colliding index, and the reimplementation of
Section~\ref{subsec:repro} re-derives real-roster values only.
Section~\ref{subsec:repro} adds a second reason to read them only as
contrasts. What is unambiguous in both
games is the ablation contrast, which spans an order of magnitude and is
significant at $p < 10^{-300}$.

One caveat applies to the real-roster figures in this game. The first game is
not affected. In the congestion game the state encodes every agent's board position,
so agents do observe the consequences of one another's actions. We have no
per-pair ground truth about which of the 12 links are genuine, and plausibly
most are. The drop from $99.8\%$ of real-roster pairs under the ablation to
$12.5\%$ under transient exclusion should therefore not be read as a
sensitivity measurement. It should also be read against the $11.0\%$ the same
procedure returns on this game's null control: the two are close enough that
this game supports the calibration claim without supporting any claim that the
corrected procedure discriminates signal from noise here. That is a real limit.
Nothing about it is presentational: in the game with the residual, the corrected
procedure detects nothing above its own error rate on this roster, and the
sensitivity evidence for the recommendation in this game comes from the
titration of Section~\ref{subsec:titration}, which injects links of known
strength into these same streams and does establish it. The social dilemma,
where the corresponding roster figures are $29.6\%$ and $3.05\%$, carries the
same claim from observation, with no injection needed. We report the congestion figure for completeness, and rest the generalization claim
entirely on the null control, whose ground truth does not depend on any
assumption about the observation channel.

\begin{table}[htbp]
\centering
\caption{Null-control false-positive rate (pairs with zero true directed
  influence by construction), pooled over 100 seeds, with Wilson $95\%$
  intervals. The precondition separates a usable test from one that
  flags everything, in both games. The paired contrast is significant at
  $p < 10^{-300}$ in each (exact McNemar). It brings the rate to the nominal
  level in the one-learner public-goods control and not in the four-learner
  congestion one; Section~\ref{subsec:roster} separates the game from the
  roster and finds most of that difference is the roster.}
\label{tab:two-games}
\footnotesize
\setlength{\tabcolsep}{3pt}
\begin{tabular}{@{}lccc@{}}
\toprule
\textbf{Game} & \textbf{Ablation} & \textbf{Exclusion} & \textbf{Nominal} \\
\midrule
Public Goods (social dilemma) & 2000/2000 & 61/2000 & \\
 & $100.00\%$ & $3.05\%$ & $5\%$ \\
 & {\scriptsize [99.81, 100]} & {\scriptsize [2.38, 3.90]} & \\[3pt]
Congestion (coordination)     & 1999/2000 & 236/2000 & \\
 & $99.95\%$ & $11.80\%$ & $5\%$ \\
 & {\scriptsize [99.72, 99.99]} & {\scriptsize [10.46, 13.29]} & \\
\bottomrule
\end{tabular}
\end{table}

Table~\ref{tab:two-games} summarizes both games. The failure mode does not
track the payoff structure: the ablation produces rates of $100\%$ and
$99.95\%$ respectively, under a training schedule the two games share. The
remedy is where they differ, bringing the rate to the nominal level in the
social dilemma as configured here, with one learner, while leaving it at more
than twice nominal in the coordination game, which has four.
Section~\ref{subsec:roster} rebuilds the first on the second's roster and
finds most of the difference is the roster.

\emph{How much does the cutoff's placement matter?} The results above use round
$8{,}000$, chosen because $\epsilon$ reaches its floor of $0.02$ at round
$\lceil \ln(0.02)/\ln(0.9995) \rceil = 7{,}823$, so the analysed window is at
the floor throughout. To measure the sensitivity of the result to that choice
we repeated the null control with the cutoff at round $7{,}000$, leaving
roughly $800$ rounds of residual decay inside the window, on identical
trajectories and identical control pairs over the first 50 of the same 100
seeds, on the earlier eight-run pool of 30 pairs each. This arm is a paired
contrast between two cutoffs on identical trajectories, so the pool's seed
reuse enters both arms identically and cancels in the comparison; we read no
absolute rate from it, and its figures are not comparable with the $3.05\%$ of
Section~\ref{subsec:te-result}, which is measured on the six-run pool.
The rate does not move
appreciably: $4.73\%$ under the derived cutoff against $4.33\%$ under the early
one, with an exact McNemar test over the 28 discordant pairs giving
$p = 0.34$. The comparison moves window
length as well as position, $4{,}000$ rounds against $5{,}000$, so it bounds
the joint effect of the two without isolating placement. The
equal-length analysis above separates them.

Residual decay of this size, about a sixth of the analysed window
and confined to the last part of the anneal, over which $\epsilon$ falls from
$0.030$ to its floor of $0.02$ and then holds there, is not enough to reproduce
the failure. The window
analysis above locates the damage in the steep part of the transient. The tail
is not where it lives. The practical implication is not that placement is delicate but the
reverse: the derived cutoff follows in one line from the schedule, and small
errors on the conservative side of it are tolerable.

One reading of this is tempting and wrong. Under
the cutoff the paper uses, the congestion rate is still elevated, which invites reading
that as the transient leaking in. The equal-length analysis excludes that
reading: in the congestion game the rate does not fall as the window recedes
from the transient, and in fact rises slightly. A rate that fails to reach
nominal is therefore not by itself evidence that the cutoff was placed badly,
and the measurement above shows the converse too, since an $800$-round error in
the cutoff does not move the public-goods rate. Distinguishing the two
requires comparing windows, which takes one extra run.

\subsection{Is It the Game, or the Roster Inside It?}
\label{subsec:roster}

The two null controls differ in two ways at once, and only one of them is the
game. Each public-goods run pairs one Q-learner with a fixed defector; each
congestion run has four Q-learners. Attributing the difference between their
post-exclusion rates to the payoff structure is therefore not safe, so we
measured it. Repeating the public-goods null control
with four learners and changing nothing else, the same $12{,}000$ rounds, the
same schedule, the same six independent runs per replication, the same 20
ordered pairs, the same 200 surrogates, the same post-transient window and the
same 100 replications, takes the rate from $3.05\%$ to $8.20\%$
($164$ of $2{,}000$; Fisher $p = 1 \times 10^{-12}$, and Mann-Whitney over the
100 seed-level rates $p = 2 \times 10^{-8}$). The apparent gap between the
games is $8.75$ points. At a matched roster it is $3.60$. Most of what looked
like the game is the roster. A game difference survives and is real (Fisher
$p = 2 \times 10^{-4}$ at the pair level, Mann-Whitney $p = 0.003$ over the 100
seed-level rates), so the payoff structure is not irrelevant, but it is the
smaller part.

Two consequences follow and both go against us.

The first is that transient exclusion does not reach the size of the test in
the social dilemma either. The $3.05\%$ reported above belongs to the
one-learner roster. The game is not what produces it: at four learners the rate is above
the size of the test at $p = 1 \times 10^{-9}$. What we had identified as the
game where exclusion suffices is the roster where it suffices.

The second is a boundary on our own recommendation. Measuring it only on the
roster where exclusion already worked would repeat the gap this subsection
exists to close, so we measured it where it is needed.
Table~\ref{tab:roster} reports all three remedies on all three null controls.
On the four-learner public-goods control none of them reaches the size of the
test, and exclusion is the lowest of the three there, $8.20\%$ against our
$9.80\%$. The difference between those two does not reach significance
(Wilcoxon $p = 0.15$ over the 100 seed-level rates), and conditioning is
decisively worse than either ($p = 3 \times 10^{-10}$). One thing survives.
The within-block surrogate is the only one of the three at or below the size of
the test in both games, $5.25\%$ and $5.50\%$ against exclusion's $3.05\%$ and
$11.80\%$ and conditioning's $5.05\%$ and $8.55\%$.
Exclusion's one-learner figure sits below the size rather than above it, which
Section~\ref{subsec:te-result} attributes to the estimator at this marginal, and
at seed level it is the lower of the two on that control, 46 seeds against 20
with 34 tied (Wilcoxon $p = 3 \times 10^{-4}$); the single-run control of
Section~\ref{subsec:withinrun} is a fourth configuration, outside this table,
and exclusion is at the size of the test there as well. What the roster costs each remedy, $4.55$ points for ours
against $5.15$ and $15.20$, separates conditioning from the other two and
leaves ours and exclusion where the paired test above left them.

\begin{table}[htbp]
\caption{Every remedy on every null control, $2{,}000$ pairs each, 100
  replications. The last row names the remedies in each column that an exact
  binomial test against the realized size of the test, $10/201 = 4.98\%$, does
  not place above it; exclusion's one-learner figure is the only cell in the
  table that falls below that size rather than above it, and it is named
  separately for that reason. The third column is the
  measurement this subsection adds: the public-goods control rebuilt with the
  congestion game's roster. Block conditioning is the conditional statistic
  defined in Section~\ref{subsec:conditional}.}
\label{tab:roster}
\centering
\begin{tabular}{@{}lccc@{}}
\toprule
 & \textbf{PGG} & \textbf{PGG} & \textbf{Congestion} \\
\textbf{Remedy} & \textbf{1 learner} & \textbf{4 learners} & \textbf{4 learners} \\
\midrule
Transient exclusion    & $3.05\%$ & $8.20\%$  & $11.80\%$ \\
Within-block surrogate & $5.25\%$ & $9.80\%$  & $5.50\%$ \\
Block conditioning     & $5.05\%$ & $20.25\%$ & $8.55\%$ \\
\midrule
At or below the size of test & all three & none & within-block \\
\bottomrule
\end{tabular}
\end{table}

Formal stationarity moves with the roster in the same way. The KPSS test,
whose null hypothesis is stationarity and which Section~\ref{subsec:stationarity}
describes in full, calls $95.7\%$ of one-learner post-transient public-goods series
stationary against $43.7\%$ at four learners
($p = 3 \times 10^{-96}$), so the sharp separation between the games that
Section~\ref{subsec:stationarity} reports is also largely roster. One thing this does not support, and we had
implied it: the stationary fraction does not order the arms. Four-learner
public goods is less stationary than the congestion game, $43.7\%$ against
$56.8\%$, and has the lower false-positive rate. The seed-by-seed correlation
within the congestion game stands as reported; reading the fraction as a
predictor across arms does not.

\subsection{A Null Control Inside a Single Run}
\label{subsec:withinrun}

Every null-control rate reported so far comes from pairs drawn across separate
runs, which is what makes their ground truth unimpeachable and is not the
situation a practitioner is in. They have one run, in which trend leakage and genuine
mediation are present together. The roster of Section~\ref{subsec:design}
cannot supply the matching control: its learners are connected through the
public signal, so no pair among them is known to be unconnected, and its
constant strategies are unflaggable whatever the truth, since permuting a
constant series returns that same series and every surrogate reproduces the
observed statistic exactly.

The control needs an agent with three properties at once. It must ignore
observations, so that influence into it is zero as a property of the code. No
assumption about the game is needed. It must carry enough entropy that the
surrogate distribution does not collapse. And it must drift, since the failure
needs a trend in both series and an agent with a flat marginal would pass the
test for the wrong reason. We add one: its action is drawn from its own
generator and the observation it is handed is discarded, while its cooperation
probability anneals on the schedule the learners' exploration follows. Its mean
marginal entropy is $0.468$ bits, against $0.911$ for a learner and $0.000$ for
the constant strategy that made the original roster unusable.

Three drifters and two learners share one match, over 100 seeds and 600 directed
pairs per group. Between drifters, where influence is zero in both directions,
the ablation flags $100.00\%$; excluding the transient gives $3.50\%$ and the
within-block surrogate $6.67\%$, neither distinguishable from the size of the
test ($p = 0.11$ and $p = 0.060$). From a learner into a drifter, also zero by
construction, the ablation flags $93.83\%$ against $5.67\%$ and $5.50\%$
($p = 0.45$ and $p = 0.51$). The failure therefore reproduces inside a single
run at full force, and both remedies control it there.

The reverse direction is not a null and is not scored as error. A drifter's
action enters the public signal the learners observe, so influence can flow that
way and only that way. It is detected in $57.50\%$ and $43.17\%$ of those pairs
against about $5\%$ in the reverse on the same pairs, which is a sensitivity
measurement with unambiguous directional ground truth. We report it because the
positive control of Section~\ref{subsec:te-result} cannot give one: its links
run through a signal that is a common input to both endpoints as well as the
path between them, so a test firing on shared input scores there as a hit. Here
the asymmetry is built into the agent.

This is the one place in the paper where the two remedies order the other way,
and the reason is in what carries the link. Neither procedure is at fault. The
titration of Section~\ref{subsec:titration} injects a lag-1 copy, so its link
lives at the fastest timescale the estimator resolves, and there the within-block
surrogate detects more at every coupling below the ceiling, and the two are
tied at $100\%$ above it. The link here is a drifter's
action entering the public signal, and a drifter's action series is a slow
deterministic ramp by construction, so the dependence it carries sits at the
same timescale as the drift. Permuting the source within blocks destroys order
inside a block and leaves the block-to-block profile intact, which is exactly
what makes the construction immune to a shared trend; the same property means a
genuine dependence carried by that trend is absorbed into the null, where it
goes uncounted. Exclusion keeps it, and detects $57.50\%$ against $43.17\%$. The two
measurements are therefore not competing estimates of one quantity. Where the
sensitivity claim is made in this paper it refers to the titration, which fixes
the coupling, matches the size and runs both arms on identical pairs; a
practitioner whose expected influence is itself slow should read this
measurement instead, and prefer exclusion.

\subsection{Can a Different Surrogate Fix It Instead?}
\label{subsec:surrogate}

A natural response to the ablation result is to blame the surrogate
construction. On that reading the data is sound. Random permutation destroys the source series'
temporal structure completely, so one might expect a circular time shift,
which preserves that structure and breaks only the alignment between the two
series, to be immune to a shared trend. If so, practitioners could keep the
convenient full-length series and simply change the null model.

We tested this directly, on the null control of both games, where the correct
answer is zero. Table~\ref{tab:surrogate} reports the outcome: the two
constructions are near-indistinguishable. With the transient included, both
flag every one of the 2{,}000 public-goods pairs, and $99.9\%$ against
$98.2\%$ of the congestion pairs. Changing the surrogate changes almost
nothing.

\begin{table}[htbp]
\centering
\caption{Null-control false-positive rate by surrogate construction, pooled
  over 100 seeds, 20 ordered pairs per seed from the six-run pool, the same
  design as Table~\ref{tab:two-games}. Changing the null model at the
  resolution of the whole series does not help. Removing the non-stationarity
  does, and so does changing the null model within blocks of training time
  (Section~\ref{subsec:conditional}).}
\label{tab:surrogate}
\footnotesize
\setlength{\tabcolsep}{4pt}
\begin{tabular}{@{}llc@{}}
\toprule
\textbf{Surrogate} & \textbf{Window} & \textbf{False positives} \\
\midrule
\multicolumn{3}{@{}l}{\emph{Public Goods}} \\
Random permutation & full series      & 2000/2000 \\
Circular time shift & full series     & 2000/2000 \\
Random permutation & post-transient   & 61/2000 \\[3pt]
\multicolumn{3}{@{}l}{\emph{Congestion}} \\
Random permutation & full series      & 1998/2000 \\
Circular time shift & full series     & 1964/2000 \\
Random permutation & post-transient   & 239/2000 \\
\bottomrule
\end{tabular}
\end{table}

The reason is that the mechanism is not what the construction assumes. The
problem is not that permutation discards the source's internal
temporal structure. It is that both series drift, so the observed pair carries
a dependence that any surrogate destroys when it breaks the correspondence
between them, whether by shuffling or by shifting. The observed statistic is
compared against a null from which the drift has been removed, and so it looks
anomalous regardless of how the removal is performed. Note that this does not
require the two drifts to be synchronized, a point
Section~\ref{subsec:schedule} establishes directly.

This is a negative result and we report it as one, with its scope stated
precisely. We tested the two constructions a practitioner is most likely to
reach for. The surrogate literature as a whole is not under test, and it is
considerably richer: the classical framework of Theiler et
al.~\cite{theiler1992} and the survey of Schreiber and
Schmitz~\cite{schreiber2000surrogate} develop amplitude-adjusted and
phase-randomized Fourier surrogates, iterative refinements of them, and
constrained-realization schemes, and trend-adjusted and model-based
constructions exist alongside these. What our result establishes is narrower than
it first appears, and we supply the counterexample ourselves in
Section~\ref{subsec:conditional}: a null model that permutes the source
\emph{within blocks of training time}
also breaks the correspondence between the series, and it does not fail. It reaches the
nominal region in both games. So the lesson is not that breaking the correspondence is
hopeless. It is that a surrogate must break the correspondence at a resolution where the
drift is flat, and both constructions tested here break it across the whole series, where
it is not. A null model that would help must therefore respect the drift
and not ignore it, which is not the standard practice in this setting and
is the open problem Section~\ref{sec:discussion} states and then narrows to
the single-run case. Reporting
which surrogate was used remains good practice, but on these data the choice
did not rescue an analysis run over a training transient.

\subsection{Manipulating the Schedule Directly}
\label{subsec:schedule}

Everything to this point identifies the training transient by elimination and
by replication across games. Neither is a manipulation. If the exploration
schedule is what drives the inflation, then changing the schedule while
holding the game, the estimator and the surrogate fixed must change the rate.
We ran that test on the null control, over 20 seeds and 400 directed pairs per
cell, in four conditions: the shared decay used throughout the paper; a
\emph{staggered} condition in which every stream uses the same decay rate but
starts from a different $\epsilon$, so the floors are reached at different
rounds; a \emph{heterogeneous} condition in which each stream draws its own
decay rate; and a \emph{constant} condition with no decay at all, exploration
pinned at the floor from the first round. Each condition was analysed both on
the full series and on the post-transient window.

Table~\ref{tab:schedule} reports the outcome, and it is only partly what we
expected. Removing the transient entirely, the constant condition, cuts the
full-series rate from $100\%$ to $23.5\%$. That is the manipulation the
mechanism predicts and it succeeds. The annealing transient is a major source.
An intervention establishes that here, where inference alone would only suggest
it, though the residual $23.5\%$ shows it is not the only one.

\begin{table}[htbp]
\centering
\caption{Null-control false-positive rate under direct manipulation of the
  exploration schedule (Public Goods; 20 seeds and 400 directed pairs per
  tabular cell, 10 seeds and 200 pairs per DQN cell). Desynchronizing the
  schedules does not help. Removing the annealing altogether does, and
  excluding the transient works in every condition in which it was run. The
  constant-$\epsilon$ DQN cell was not run post-transient. The text below
  says why that is the gap in this arm we would close first.}
\label{tab:schedule}
\footnotesize
\setlength{\tabcolsep}{5pt}
\begin{tabular}{@{}lcc@{}}
\toprule
\textbf{Schedule} & \textbf{Full series} & \textbf{Post-transient} \\
\midrule
\multicolumn{3}{@{}l}{\emph{Tabular Q-learning}} \\
Shared decay (default)        & $100\%$   & $2.2\%$ \\
Staggered start               & $100\%$   & $4.3\%$ \\
Heterogeneous decay rates     & $100\%$   & $2.8\%$ \\
Constant $\epsilon$, no decay & $23.5\%$  & $3.3\%$ \\[3pt]
\multicolumn{3}{@{}l}{\emph{DQN}} \\
Shared decay                  & $100\%$   & $1.5\%$ \\
Constant $\epsilon$, no decay & $44.0\%$  & not run \\
\bottomrule
\end{tabular}
\end{table}

The other three conditions correct us. We had described the mechanism as
agents whose action probabilities move \emph{in step}, which implies that
desynchronizing the schedules should help. It does not. Staggering the
starting points leaves the rate at $100\%$, and so does giving every agent its
own decay rate. What matters is that both marginals move over the run.
Whether those paths are aligned in time or share a rate makes no difference, and
Section~\ref{subsec:stationarity} reproduces the failure with a non-monotone path,
so monotonicity is no part of the condition either. This result matters
practically, because the natural mitigation available to a practitioner,
varying the exploration schedule across agents so the runs are not
synchronized, does not reduce the false-positive rate.

The residual $23.5\%$ under constant exploration is also informative. With the
annealing removed there is no exploration transient, yet the full series still
produces false positives at more than four times the nominal rate, while the
post-transient window of the same runs sits at $3.3\%$. The annealing schedule
is not the only thing that leaves a trend in the early part of a run. Learning
itself does it too. That
is the same phenomenon Section~\ref{subsec:stationarity} identifies behind the
coordination-game residual, appearing here in the simpler setting.

\paragraph{The effect is not an artefact of tabular learning} Every result so
far uses tabular Q-learning, which invites the objection that the inflation is
a property of table-based updates. On that objection, learning under annealed
exploration is not the cause. We therefore repeated two cells with DQN agents, which replace
the table with a neural function approximator and add a replay buffer and
periodic target-network updates, each an independent source of
non-stationarity. Nothing else changed: same game, same estimator, same
surrogate, same null control.

The outcome is identical where it matters and worse where the diagnosis says
it should be. Under the shared schedule the DQN full series produces $200$
false positives out of $200$, exactly as the tabular agents do, and excluding
the transient brings that to $1.5\%$, below nominal on 200 pairs (Wilson
$[0.5\%, 4.3\%]$, $p = 0.021$; the seed-level interval over the ten seeds is
$[0.0\%, 3.2\%]$, so the two levels agree). Neither the
failure nor the remedy depends on the architecture. Under constant
exploration, however, the DQN residual is $44.0\%$ against the tabular
$23.5\%$, and it varies widely across seeds (seed-level standard deviation
$0.26$, with one seed at $100\%$). This is consistent with additional sources
of learning-induced non-stationarity in the DQN agents. A DQN carries sources
of non-stationarity a table does not,
so once the exploration transient is removed, what is left to detect
increases. It does not fall. The implication runs the wrong way for deep
multi-agent work. The more machinery an agent carries, the less of the problem
the exploration schedule accounts for, and the more a stationarity check is
needed in place of a schedule-derived cutoff. The cell that would sharpen this,
DQN under constant exploration with a window excluded, we did not run, and it
is the gap in this arm we would close first: with no annealing there is no
schedule-derived cutoff, so what a stationarity-selected window recovers from
a $44.0\%$ residual is the quantity the recommendation turns on for deep
agents.

\paragraph{Conditioning on the target's own past does not help} Transfer
entropy is a conditional mutual information, $I(X_{t+1}; Y_t \mid X_t)$, so it
already conditions on the target's own history. That conditioning
distinguishes it from plain MI, and one might expect it to absorb a shared
trend. It does not. Repeating the null control with pairwise mutual
information between the same independently-generated series, same estimator,
same 200 permutation surrogates, same two windows, gives $400$ of $400$ pairs
flagged on the full series against $2.2\%$ after the transient is excluded:
indistinguishable from transfer entropy. The post-transient counts are
identical, $9$ of $400$ each, and that is a coincidence. The two statistics are
not duplicating each other: they run on the same 400 pairs and flag almost disjoint sets,
sharing only one pair between them, as two tests near the same rate on null
data should. These are nested members of one
family. They are not two independent measures, and the point of running both is
precisely that the conditioning already built into TE gives no protection here.
If both action probabilities drift with training time, then pooling over time
makes the series appear dependent, because knowing where one sits in its own
trajectory locates the other. That is a property of the pooling. The agents are
not the source of it. We report two measures where four were available, because
entropy and KL divergence are point estimates and never produce a significance
decision, so neither has a false-positive rate to measure, and the design used
throughout this section applies only to quantities that do.

\subsection{A Remedy That Keeps the Data}
\label{subsec:conditional}

Excluding the transient works, but it discards two thirds of every run. That
trade-off deserves examination, because the framing that motivates it is
incomplete. We have been treating the drift as something to be removed from
the data. It is equally a variable that can be conditioned on, and in this
setting it is an unusually convenient one: the exploration schedule is chosen
by the experimenter, is deterministic, and is known in closed form. The drift
is not a latent confounder. It is an observed one, and the standard response
to an observed common driver is to put it in the conditioning set. That route
reaches the same place without a new null model, and
Section~\ref{subsec:titration} compares the two against injected links.

For transfer entropy that means computing
\begin{equation}
T_{Y \to X \mid Z} = I(X_{t+1};\, Y_t \mid X_t,\, Z_t),
\end{equation}
where $Z_t$ discretizes training time. Conditioning does not make the
exploration rate constant within a block, and the conditional sweep of
Table~\ref{tab:conditional} is best read as measuring how much of it survives.
Two sweeps below vary the block count and they are not the same experiment: that
one conditions the statistic, while the sweep in the paragraph on whether the
block count is tuned leaves the statistic unconditioned and varies only the
surrogate, so their rates differ at every count and only the conditional ones
belong beside the figures that follow. The worst within-block
relative fall in $\epsilon$ is $98\%$ at one block, $77.7\%$ at four, $52.7\%$
at eight and $17.1\%$ at 32, and the false-positive rate tracks that sequence
monotonically at $100\%$, $65.0\%$, $12.0\%$ and $6.0\%$. What conditioning
buys is therefore a dose of the driver removed. It does not eliminate the
driver, and the dependence that survives is whatever the residual within-block
drift cannot explain. The
surrogate is drawn to match: the source is permuted \emph{within} each block,
which destroys the source-target correspondence while leaving the block
structure, and therefore the drift, intact. Permuting across blocks would
reintroduce exactly what the conditioning removes.

\emph{Which of the two changes does the work?} That arm changes two things at once, so
the rate alone cannot say. We ran the missing cells: the ordinary unconditional statistic
against within-block surrogates, and the conditional statistic against whole-series ones,
on the same pairs.

\begin{table}[htbp]
\centering
\caption{Separating the conditioning from the surrogate, null control on the full
  untrimmed series, 100 seeds and $2{,}000$ pairs per cell. The surrogate change alone is
  the only one of the four at the size of the test in both games; conditioning alone
  overshoots into the conservative region, and the two together are calibrated in the
  social dilemma but not in the coordination game.}
\label{tab:decomposition}
\footnotesize
\setlength{\tabcolsep}{5pt}
\begin{tabular}{@{}llcc@{}}
\toprule
\textbf{Conditioning} & \textbf{Surrogate} & \textbf{Public Goods} &
\textbf{Congestion} \\
\midrule
none & whole series   & $100.00\%$ & $99.95\%$ \\
none & within block   & $5.25\%$  & $5.50\%$ \\
32 blocks & whole series & $0.90\%$ & $2.85\%$ \\
32 blocks & within block & $5.05\%$ & $8.55\%$ \\
\bottomrule
\end{tabular}
\end{table}

Table~\ref{tab:decomposition} answers the question, and the answer is not the one the
sweep at 10 seeds suggested. The surrogate carries the correction on its own: changing
nothing but how the source is permuted takes the rate from $100.00\%$ to $5.25\%$ and
from $99.95\%$ to $5.50\%$, and neither is distinguishable from the size of the test
(exact binomial $p = 0.57$ and $p = 0.28$). Conditioning alone also removes the
inflation but overshoots it, to $0.90\%$ and $2.85\%$, both measurably below nominal
($p = 3 \times 10^{-24}$ and $p = 3 \times 10^{-6}$). That asymmetry is one we would
expect: a conditional statistic compared against a null that has already destroyed the
drift is being asked to clear too low a bar, and pays for the safety in power.

In the social dilemma the two calibrated cells are interchangeable: $5.25\%$ for the
surrogate alone against $5.05\%$ for block conditioning with within-block
surrogates, Wilcoxon signed
rank $p = 0.87$ over the 100 seed-level rates. The test is paired because the
four cells run on the same $2{,}000$ pairs, and the seed is the unit because
pairs within a seed share runs. Thirty-nine of the 100 seeds give identical
rates in the two cells, and how those ties are handled moves the figure:
discarding them gives $0.87$ and splitting them gives $0.41$. We report both
because the claim here is a null one, and neither is near rejection. In the coordination game they are not
interchangeable. There the surrogate change
alone is the only one of the four at the size of the test, $5.50\%$ against the conditioned
combination's $8.55\%$, which remains above nominal ($p = 2 \times 10^{-11}$); the
difference between the two is itself significant (Wilcoxon $p = 5 \times 10^{-4}$). The
remedy that game needs is therefore the simpler of the two: permute within blocks and
leave the statistic alone.

Every cell above is measured at $2{,}000$ pairs. The sweeps that follow are not,
and we read them as orders of magnitude. They are not calibration figures, for
a reason this table demonstrates: a 200-pair estimate of one of these cells moves
by a factor of two and a half at the larger sample without the conclusion
changing.

\emph{Is the block count tuned?} The construction has one free parameter and it
was not chosen for it: 32 comes from the conditional sweep below, run on a
different statistic in one game against this same null control. If the rate
crossed the size of the test steeply there, the number would be a value read
off where a curve happened to cross. Swept over the whole realizable range on
the null control, it does not. One and two blocks reproduce the $100\%$ failure,
since permuting within one block is permuting the whole series; four blocks give
$78.0\%$ and eight $16.5\%$; and then 16, 32, 64, 128 and 256 blocks give
$5.5\%$, $2.0\%$, $2.0\%$, $3.0\%$ and $3.0\%$, none of them distinguishable
from the size of the test. The collapse is therefore finished by 16 blocks, and
every setting from there to 256, a sixteenfold range, lands in the nominal
region. We state that as the collapse completing early. Calling it a flat rate
would overstate it, because these are 200-pair cells: the 32-block cell reads $2.0\%$ here and
$5.25\%$ when the same construction is measured at $2{,}000$ pairs in
Table~\ref{tab:decomposition}, and the others were not re-measured. What the
sweep establishes is that the choice is not critical, which is the property a
practitioner who has not run it needs; what it does not establish is equality
among the settings inside the region.

The parameter can also be removed outright. The partition that splits
recursively until KPSS accepts each segment, described below, has no free
parameter; applied to the within-block surrogate it selects a median of 13
blocks and returns $2.5\%$, again at the size of the test. Removing the
parameter is attractive for that reason, but calibration is not the only
criterion and the sensitivity measurement below decides against it; the fixed
count is what we recommend, and what the rest of this section uses. As with the
conditional sweep, the 32-block cell
here is the first ten seeds of the 2{,}000-pair measurement in
Table~\ref{tab:decomposition}, $4/200$ contained in $105/2{,}000$.

Table~\ref{tab:conditional} reports the null-control false-positive rate of
this statistic on the full, untrimmed 12{,}000-round series, swept over the
number of blocks. The sweep runs at 10 seeds, where the calibration cells run
at 100, because it is a sweep: each of its six cells costs a full surrogate test on every pair, and
what it has to establish is the shape of the curve from $100\%$ down to the
nominal region, which 200 pairs per cell resolve comfortably. The endpoint
that matters is the claim that conditioning is calibrated, which is a stronger
statement than being merely better. It is not settled by this table alone but
by its agreement with the adaptive partition below and with the sensitivity
check on the real roster. The unconditional statistic is the one-block case, and it
reproduces the $100\%$ failure. Conditioning on two blocks changes nothing,
four blocks reduces the rate to $65\%$, and by 32 blocks of 375 rounds each
the rate is $6.0\%$, on data from which nothing has been discarded. At 200
pairs that is not separable from the nominal $5\%$, Wilson $[3.5\%, 10.2\%]$,
so what the sweep establishes is the collapse to the nominal region. It does
not establish a calibration figure. The calibration figure for its endpoint is measured
separately at 100 seeds in Table~\ref{tab:decomposition}, where the same
32-block partition gives $5.05\%$ on $2{,}000$ pairs; these ten seeds are the
first ten of that measurement and its $12/200$ is contained in its $101/2{,}000$.

\begin{table}[htbp]
\centering
\caption{Null-control false-positive rate of block-conditioned transfer
  entropy on the \emph{full} series (Public Goods, 10 seeds, 200 directed
  pairs per cell). One block is the unconditional statistic. The final column
  is the count of genuinely connected pairs recovered on the real roster, out
  of 60, over the same seeds.}
\label{tab:conditional}
\footnotesize
\setlength{\tabcolsep}{5pt}
\begin{tabular}{@{}lccc@{}}
\toprule
\textbf{Blocks} & \textbf{Rounds each} & \textbf{False positives} &
\textbf{True links} \\
\midrule
1 (unconditional) & 12{,}000 & $100\%$ & 60/60 \\
2                 & 6{,}000  & $100\%$ & \\
4                 & 3{,}000  & $65\%$  & \\
8                 & 1{,}500  & $12\%$  & 60/60 \\
16                & 750      & $8.5\%$ & \\
32                & 375      & $6.0\%$ & 60/60 \\
\bottomrule
\end{tabular}
\end{table}

The sensitivity check matters more than the calibration here, because a
statistic that has simply stopped responding would show the same collapse in
false positives. It has not. On the real roster, analysed on the full series
at 8 and at 32 blocks, the conditional statistic recovers all $60$ of the $60$
genuinely connected pairs across the ten seeds. Conditioning removes the drift
and leaves the signal. The remaining pairs on that roster carry a constant
strategy at one endpoint and are excluded here for the reason given in
Section~\ref{subsec:te-result}.

\emph{The remedy does not transfer to the control that needed it.} Everything
above is measured on the one-learner social dilemma, where excluding the
transient had already reached nominal. The coordination game is the one with an unexplained
residual, so it is the one on which a remedy has to be shown, and we ran the
identical sweep there, changing only the game. Conditioning removes most of the
inflation and does not close it: the full series falls from $99.95\%$ to
$17.0\%$ at eight blocks and $8.55\%$ at 32, still above the size of the test
where the same partition reaches $5.05\%$ and
nominal in the social dilemma. The 32-block figures here are the $2{,}000$-pair
measurements of Table~\ref{tab:decomposition}, and at that sample the shortfall
is unambiguous (exact binomial $p = 2 \times 10^{-11}$).
On the post-transient window it moves the
residual only from $11.5\%$ to $9.5\%$.

That is what the diagnosis predicts, so we state it as a boundary. It is not
a disappointment. Conditioning on $Z$ removes the dependence $Z$ creates, and
in this game a substantial part of the drift is not the schedule:
Section~\ref{subsec:stationarity} finds $43.2\%$ of series still non-stationary
after the transient is gone, and Section~\ref{subsec:schedule} finds a $23.5\%$
residual under constant exploration. Conditioning therefore handles drift with
a known source and not the coordination residual, which is narrower than we
could state before running this. The remedy that does reach the size of the
test in that game is the within-block surrogate measured earlier in this
subsection.

\emph{Conditioning on the other series does not help either.} Everything above is
bivariate, and the obvious objection is that the failure might be an artefact of
that: if every run in the pool anneals on the same schedule, then conditioning
the statistic on a third run's series might absorb the shared component and the
spurious dependence could disappear without any remedy at all. That is the
correction a practitioner with a multivariate toolbox would reach for first, and
if it worked the result here would belong to the estimator. The surrogate
procedure would be in the clear. On the null control, full untrimmed series, nothing
discarded and only the conditioning set changing, it does not work at all.
Against $2{,}000$ of $2{,}000$ for the bivariate statistic, conditioning on one
other run gives $2{,}000$ of $2{,}000$ and conditioning on the joint state of two
others gives $2{,}000$ of $2{,}000$. Not one pair escapes in any arm.

That is what the mechanism predicts, and it is no surprise. The surrogate
destroys the drift in the source, so the observed statistic is compared against a
null that lacks a feature the data has; adding more drifting series to the
conditioning set changes what is being estimated but not what the comparison is
against. Conditioning helps only when the partition it induces is coarse enough
in time to leave the drift inside each cell, which the block index does
and what another agent's action series does not.

Two further limits should be stated. First, conditioning works because the driver is known
and observed; where the drift arises from something the experimenter cannot write down it
is unavailable, and the within-block surrogate, which needs no conditioning set, or a
trend-preserving construction of the kind developed by Lucio et al.~\cite{lucio2012},
is what remains. Second, where blocks are scarce the partition has to put its resolution
where the driver moves: binning by equal intervals of $\epsilon$, in place of
equal intervals of time, cuts the four-block rate from $65\%$ to $43\%$, while equal \emph{counts} of $\epsilon$
return $66.5\%$, since a third of the run sits at the exploration floor. The schemes
converge once blocks are plentiful, and an irregular schedule would make the choice matter
more.

\paragraph{Removing the free parameter} The block count above was swept.
Nothing derived it, which is unsatisfying in a paper whose argument is that
preconditions should be checked, never assumed. The stopping rule is
available from the diagnostic reported in Section~\ref{subsec:stationarity},
the KPSS test, whose null hypothesis is that the series is stationary. Split the
series recursively, halving whenever KPSS rejects stationarity for the current
segment and stopping when it accepts or when a further halving would fall
below the length the plug-in estimator can support. The partition is then a
property of the data. Applied to the null control it chooses a median of 13
blocks, ranging from 9 to 18 across pairs, and produces unequal blocks: fine
early where $\epsilon$ falls quickly, coarse over the floor where nothing
changes. The resulting false-positive rate is $6.5\%$, matching the $6.0\%$
the same ten seeds give when the count is hand-tuned to 32, with no count to tune.

That comparison is 200 pairs and says nothing about sensitivity, so we ran both
arms at the power the rest of the section uses and then titrated them. On
$2{,}000$ pairs of the null control the derived partition returns $109/2{,}000 =
5.45\%$ against $103/2{,}000 = 5.15\%$ for the fixed count, neither separable
from the size of the test ($p = 0.33$ and $p = 0.72$) nor from each other: the
two run on identical pairs, so the contrast is paired, and exact McNemar on the
54 and 48 discordant pairs gives $p = 0.62$. The fixed arm here is the same
procedure as the $5.05\%$ cell of Table~\ref{tab:decomposition} on the same
pairs, differing only in the surrogate draw. Calibration therefore does not
choose between them, and the median partition of 13 blocks at $2{,}000$ pairs
ranges from 7 to 21.
Sensitivity does choose, against the derived partition. Injecting the same
graded links into the same pairs, the two are indistinguishable on the null row
and at the two strongest couplings, but at $c = 0.02$ the derived partition
detects $14.0\%$ against the fixed count's $21.0\%$, losing 19 pairs the fixed
count finds and gaining 5 (exact McNemar $p = 0.007$, and Wilcoxon
$p = 0.008$ over the ten seed-level rates). Removing the parameter
therefore costs power exactly where power is scarce, which is the regime the
titration in Section~\ref{subsec:titration} identifies as the one that matters.
We report the derived partition as a way to remove a knob. It is not an
improvement, and we do not recommend it over the fixed count.

We also tested the alternative of keeping every block but trusting them
unequally, estimating transfer entropy within each block, bootstrapping its
variance there, and combining with inverse-variance weights. The motivation is
that conditioning discards all between-block information, while weighting only
discounts it. On these data it does not pay: the weighted statistic reaches
$9.0\%$ against $6.5\%$ for conditioning on the same partition. What that
supports is the absence of the predicted improvement. Nothing here reverses it. The
two rates come from 200 pairs, where the paper's own standard applies as much
here as anywhere: they are not separable from one another (Fisher exact
$p = 0.45$), so we do not claim that weighting is worse, only that it is not
better. One caveat on that test, since it runs the other way from our
conclusion. The two arms are computed on the same pairs, so the comparison is
paired and a paired test would be the more powerful one; this arm did not retain
per-pair outcomes, so we cannot recompute it, and an unpaired test is the easier
route to a finding of no difference. The honest reading is that we did not
establish a difference here. Whether one exists is open. The mechanism we would expect, that downweighting a block leaves the
drift surviving inside it in the pooled statistic while conditioning removes
the comparisons across blocks outright, is consistent with the numbers and is
not established by them. We report the arm because it is a natural alternative
and its failure to improve on conditioning is informative.

That refutation should be read at the resolution at which we tested it. The
weights are coarse: one scalar per block, from a 60-resample bootstrap of that
block's own transfer entropy, with no pooling across blocks and no model of
how the variance ought to behave as a function of block length or of the
marginal entropies inside it. A finer scheme, shrinking the per-block
variances toward a common estimate, or deriving them from the known
finite-sample behaviour of the plug-in estimator where we resampled, might
close the gap or reverse it. What the result establishes is that weighting as
naively implemented does not beat conditioning. Some other weighting might. Neither limit affects the main result of this paper,
which concerns what happens when the precondition is ignored, but both bear on
how the remedy should be used.

\paragraph{Continuous actions and a kNN estimator} Binary actions and a
discrete plug-in estimator run through everything above, which leaves open
whether the inflation is an artefact of discretization. We therefore repeated
the null control in a continuous-action public goods game, where each agent
contributes $c \in [0,1]$, samples from a Gaussian policy whose standard
deviation anneals in place of $\epsilon$, and updates its mean by a
REINFORCE step. Transfer entropy is estimated by the conditional
Kraskov-St\"ogbauer-Grassberger scheme with $k = 4$ under the max norm, so no
binning enters at any point.

The failure survives the change intact. On the full series, $39$ of $40$
control pairs are flagged, $97.5\%$, where the discrete estimator gives
$100\%$. The mechanism is not about discretization.

The remedies transfer unevenly, on an arm sized to resolve a large effect. A
small one is beyond its resolution. Excluding the
transient brings the rate to $12.5\%$ and block conditioning to $22.5\%$, both
far below the unconditional rate but both above the size of the test (Wilson
$[5.5\%, 26.1\%]$ and $[12.3\%, 37.5\%]$, exact binomial $p = 0.047$ and
$p = 1 \times 10^{-4}$). Only the second survives at seed level: over the ten
seeds the intervals are $[0.0\%, 25.1\%]$ and $[6.8\%, 38.2\%]$, and the first
contains the size of the test. Forty pairs are too few to place a rate of
$12.5\%$, which is the resolution limit this arm was built at. The within-block surrogate, which is what we recommend,
is the one of the three that is not: $4$ of $40$, $10.0\%$, $[4.0\%, 23.1\%]$,
$p = 0.14$. That ordering holds at four learners, in the congestion game and
here, and reverses nominally only at one learner in the social dilemma, where
the two are interchangeable. This arm cannot establish it either: $10.0\%$ against $22.5\%$ on the same 40 pairs
is Wilcoxon $p = 0.25$ over the ten seed-level rates, so what this shows is that the recommendation is the only one
of the three not distinguishable from nominal here. It does not show that it
beats the others here. Conditioning also reads worse than exclusion, $22.5\%$ against $12.5\%$,
and that ordering is equally unestablished at this sample. Two
explanations are available and our design does not separate them. The
post-transient window in this arm is $1{,}500$ samples, where a kNN estimator
in three dimensions is working close to its limit, so part of the residual may
be estimator variance. Surviving drift need not explain it. Conditioning does
not help here either, and makes matters worse, since partitioning into 16
blocks leaves roughly $280$ points per block for a neighbour-based estimator. The honest reading is
that the diagnosis generalizes to continuous data while the remedies, as
implemented here, are sample-hungry in a way the discrete versions are not.

\subsection{Estimating Across Runs Instead of Over Time}
\label{subsec:ensemble}

Every remedy so far treats one series and inherits the assumption that makes the problem:
that samples from round 100 and round 11{,}900 may be pooled. Ensemble estimators, developed
for neuroscience where trials are the natural unit and stationarity within a trial cannot be
assumed~\cite{wollstadt2014}, do not. They are used for exactly this reason in practice:
studies of directed coupling between brain areas during learning tasks adopt the ensemble
over repeated trials. A single trial is not stationary, which is the ground they
state~\cite{zhu2023}. We take $E$ independently seeded runs and, at each
round, estimate the conditional mutual information from the $E$ samples available there, all
drawn from the same distribution whatever that distribution is, then average over rounds.
Non-stationarity across time becomes irrelevant by construction, with no preprocessing.
The surrogate permutes which run supplies the source, so every series keeps its own trend
while the correspondence between runs is destroyed.

On the full untrimmed series the estimator returns $1$ of $30$ pairs in the social dilemma at
$E = 50$ and $2$ of $30$ in the coordination game at $E = 30$, $3.3\%$ and $6.7\%$, against
$100\%$ and $99.3\%$ for the pooled statistic on the same runs. Neither departs from the
size of the test. But 30 tests are 30 tests: the congestion interval is $[1.8\%, 21.3\%]$,
which contains both the $11.8\%$ exclusion achieves and the $8.55\%$ conditioning achieves,
so we cannot claim an improvement on either and do not. Section~\ref{subsec:conditional}
refuses an ordering between $9.0\%$ and $6.5\%$ on 200 pairs, and 30 tests carry far less
power than that.

What the arm establishes is that a construction which never pools over time is not defeated by
the drift that defeats pooling, across the whole range from $100\%$ to nominal and in both
games. That is a demonstration of principle. Widening the interval is expensive. The
obstacle is compute, never labour: an ensemble of $E$ runs buys one test where $E$ runs
analysed singly would buy $E$, so these 30 tests already cost more compute than the 200
pairs of any other arm. The per-round estimate over eight cells
is also heavily biased, so this arm measures calibration. Effect size is beyond it. The bias
is common to the statistic and its surrogates and cancels, as with the Miller-Madow offset of
Section~\ref{subsec:other-measures}. For the coordination game the remedy we can recommend
on measured grounds remains the within-block surrogate above, at $5.50\%$ on $2{,}000$ pairs.

\subsection{Why Not Simply Difference the Series?}
\label{subsec:detrend}

Removing a trend before estimating is the first move anyone working with time
series would make, and it deserves a measurement. On
the same null control, at ten seeds and 200 pairs from the six-run pool, the
untreated full series flags $100\%$, excluding the transient gives $5.5\%$,
first differences give $5.0\%$ and block conditioning at 32 blocks gives
$6.0\%$. The conditioned figure reproduces Table~\ref{tab:conditional}
exactly, which is the internal check that these are like for like, and the
exclusion figure is the 10-seed counterpart discussed above.

Differencing therefore controls the rate as well as anything we recommend, and
the reason to prefer conditioning is not that it fails. Both change the
estimand, and it is worth being exact about how, since the same objection
applies to our own remedy. Differencing changes the variables: it estimates
$T(\Delta Y \to \Delta X)$, a transfer entropy between two processes that are
not the ones under study. Conditioning keeps those variables and widens the
conditioning set, estimating $T(Y \to X \mid Z)$, the influence not already
accounted for by training time. Neither is $T(Y \to X)$. Of those two we
prefer the second
because its estimand is the original question asked within blocks of the
driver, which is interpretable as a claim about the agents, whereas the first
is a claim about their increments.

Why differencing helps is worth stating precisely, because the loose version
of the answer is wrong in a way that matters. It does not make the series
stationary. Measured by quartile, the differenced series shifts in
distribution \emph{more} than the original, $P(\Delta = 0)$ moving by $0.345$
across the run against $0.256$ for $P(x = 1)$. What it removes is narrower
than that: the first moment alone, whose range falls from $0.256$ to
$0.00033$. Since the false positives disappear anyway, the leakage cannot
require co-varying distributions. What it does require is a common drift in
the means, which is what the schedule manipulation of
Section~\ref{subsec:schedule} isolated by intervention. It is
a necessary condition. It is not a complete account, since that manipulation
leaves a $23.5\%$ residual under constant exploration.

One further explanation can be ruled out, because a reader will reach for it.
Differencing a binary series might appear to widen the conditioning set and
detrend nothing, since $\Delta x_t$ is determined by two consecutive
values. It cannot: the map is lossy in the direction that matters, because
$\Delta = 0$ arises from both $(0,0)$ and $(1,1)$, so the differenced series
carries strictly less than the pair it came from. Measured directly,
conditioning the ordinary statistic on two lags of the target leaves the rate
at $100\%$, so extra history is not what differencing supplies.

Detrending in the ordinary sense is not available here at all, and the reason
is worth stating because the obvious implementation silently does nothing.
Subtracting a local mean from a binary series and rethresholding at zero
returns the original series exactly: where $x = 1$ the residual $1 - m$ is
positive and where $x = 0$ the residual $-m$ is negative, so the sign pattern
is $x$ itself. On our data that transform changes $0$ of $12{,}000$ samples.
Keeping the residual continuous and estimating with the same
Kraskov-St\"ogbauer-Grassberger scheme used in the continuous-action arm does
change the data. Matters get worse, and every one of the
40 pairs is flagged. The residual takes two non-overlapping branches, $-m(t)$
and $1 - m(t)$, so its sign still carries the action while its magnitude
carries the local mean, which correlates $0.77$ to $0.80$ with position in the
run. Detrending this way encodes training time more explicitly than the binary
series did. We do not claim that no detrending scheme could work, only that the
routes a practitioner would try first either do nothing, change the estimand,
or make the leakage easier to find.

\subsection{How Strong Must a Real Link Be?}
\label{subsec:titration}

Three procedures now control the false-positive rate. None has yet been
asked whether it can still find a link that is genuinely there. A null control
establishes only that a procedure does not fire when it should not, and a
procedure that never fires would pass it perfectly. The recovery of 593 of 600
connected pairs argues against that reading, but every link in that roster is
of broadly similar strength, so the question of weak links stays open. Since
excluding the transient discards two thirds of the data, the sensitivity that
costs has to be measured.

We therefore vary the strength of a genuine link continuously. The
construction starts from the null control itself, two Q-learners trained in
separate non-interacting matches, so the shared exploration-decay trend is
present at full strength, and writes a lag-1 link of controlled size into the
target:
\begin{equation}
X_t =
\begin{cases}
Y_{t-1} & \text{with probability } c, \\
X^{0}_t & \text{otherwise,}
\end{cases}
\label{eq:coupling}
\end{equation}
with $X^{0}$ the agent's own uncoupled action. At $c = 0$ this is the null
control exactly, so the first row of the sweep is a consistency check. No new
claim rests on it. It is not an independent replication and we do not read it as
one: these ten seeds draw the same trajectories from the same builder as the
arms above, and only the surrogate implementation differs, which is why the same
cell reads $6$ of $200$ here and $4$ of $200$ in
Table~\ref{tab:decomposition}. What the row establishes is that a second
implementation of the same construction agrees. Whether the result holds on
fresh data is untouched. The four
procedures are then applied to identical data: the ablation on the full
series, transient exclusion on the post-transient tail, and the within-block
surrogate and block conditioning with 32 blocks, both on the full series.

\begin{figure}[htbp]
\centering
\includegraphics[width=\linewidth]{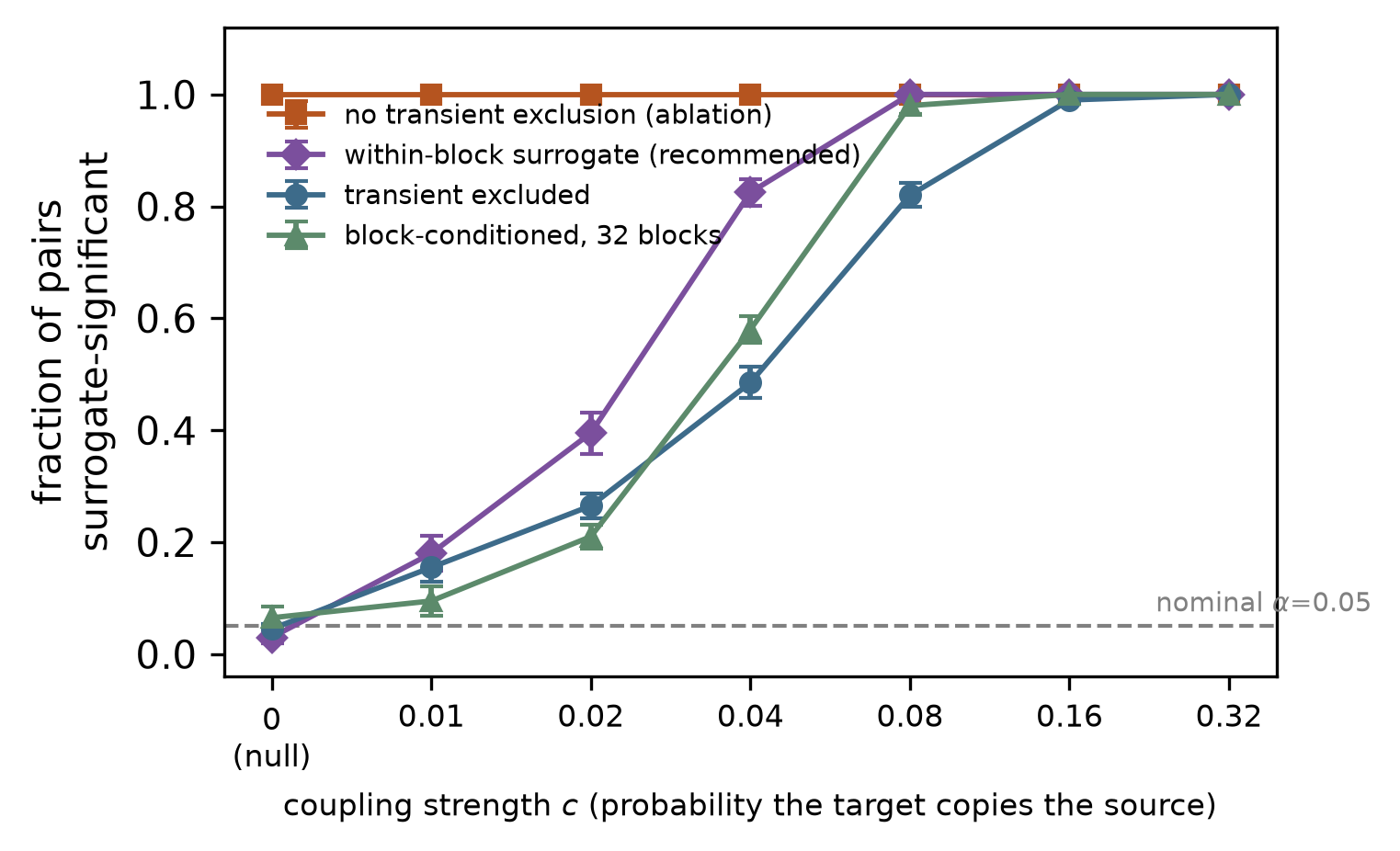}
\caption{Detection rate against the strength of a genuine lag-1 link, injected
  by Eq.~\eqref{eq:coupling} into the null-control streams so that the shared
  exploration trend is retained. Error bars are standard errors across 10
  seeds. The ablation is flat at $100\%$ across the whole range, firing no
  less often when there is nothing to detect. The three corrected procedures
  rise clearly above the nominal level from $c \approx 0.02$ onward. The
  within-block surrogate is never below either of the other two, strictly
  highest up to $c = 0.08$ and tied with conditioning at the ceiling above it,
  while starting from
  the lowest null rate, and exclusion and conditioning cross each other near
  $c = 0.03$.}
\label{fig:titration}
\end{figure}

Table~\ref{tab:titration} and Fig.~\ref{fig:titration} report the outcome. The
$c = 0$ row recovers $4.5\%$ and $6.5\%$ against the $5.5\%$ and $6.0\%$ a
separate implementation returns on the same ten seeds
(Section~\ref{subsec:detrend}), and the ablation again flags
everything.

Three things follow. First, the ablation's apparent sensitivity carries no
information whatever: it reports $100\%$ at $c = 0$, where no link exists, and
$100\%$ at $c = 0.32$, where a strong one does. A procedure whose output is
identical with and without the effect it purports to detect is not a
sensitive test but a constant function.

Second, weak genuine links are lost, and the threshold is measurable, though
it is not the same threshold for every procedure. Exclusion falls below one
half at $c \approx 0.04$, corresponding to roughly $7.6 \times 10^{-4}$ bits;
conditioning still detects $58.0\%$ there, and the within-block surrogate
$82.5\%$. At $c = 0.01$ conditioning is barely above its own false-positive
floor, $9.5\%$ against $6.5\%$, while exclusion reaches $15.5\%$ and the
within-block surrogate $18.0\%$. The threshold below which none of the three
recovers half the links is $c = 0.02$, about $3.6 \times 10^{-4}$ bits, and
within this design an influence weaker than that is not reliably recoverable
at 12{,}000 rounds. That is a real limitation on the guardrail.
The null control could not have revealed it.

Third, the ordering among the remedies is the reverse of what their cost would
suggest. Against an injected lag-1 link the construction that changes least is
the more sensitive of the two that share a window, and
Section~\ref{subsec:withinrun} reports the one case
where a link of a different temporal character reverses the comparison with
exclusion, which is the other pair and the weaker of the two contrasts. Because
all four procedures run on identical pairs the comparison is paired, and we
test it with exact McNemar. Against exclusion, the within-block surrogate finds
83 pairs that exclusion misses while missing 15 that exclusion finds at
$c = 0.04$ ($p = 1 \times 10^{-12}$), and 36 against none at $c = 0.08$
($p = 3 \times 10^{-11}$). Against conditioning it finds 60 against 11 at
$c = 0.04$ ($p = 3 \times 10^{-9}$). Each of these three contrasts spans less
than an order of magnitude, so each carries a seed-level figure beside it: the
signed-rank test over the ten seeds returns $p = 0.002$ for all three, the
smallest value ten seeds admit, so no seed goes the other way.
At $c = 0$ none of the three separates
from another on the null control, which is how we compare them at
matched size, although 200 pairs would not resolve a factor of two in realized
size and we do not claim more than that.

The two contrasts are not equally clean, and the difference is worth stating
because the weaker one is the more quotable. Within-block and conditioning run
on the same series over the same window, so their comparison isolates the
construction. Exclusion does not: it analyses the post-transient tail, where the
source's marginal entropy averages $0.081$ bits against $0.420$ over the full
series. At a fixed injection probability the link therefore carries several
times less information in the window exclusion sees, before any question of
sample size, and the shortfall H5 attributes to answering the same question from
a third of the sample has this second component beside it. Both are properties
of where the window sits. The surrogate construction is not involved. We therefore
rest the sensitivity claim on the within-block against conditioning contrast,
and report the exclusion comparison as an upper bound on the gap. It is not a
measurement of it. Conditioning
in turn beats exclusion in the mid range, 46 against 27 at $c = 0.04$
($p = 0.034$), for the reason H5 gave: exclusion answers the same question from
a third of the sample.

\emph{And in the control that needs it.} Everything above is the one-learner
social dilemma, where excluding the transient already sits at or below the size
of the test, so it is not the control a better remedy exists for. We therefore ran the identical titration in
the coordination game, changing nothing but the environment. In the null row the
within-block surrogate returns $6.0\%$, not distinguishable from the size of the
test ($p = 0.51$), while exclusion sits at $11.0\%$ ($p = 5 \times 10^{-4}$) and
conditioning at $8.5\%$ ($p = 0.03$). That pair is the one place in this
paper where the two levels of analysis disagree, and the rule of
Section~\ref{subsec:design} is to say so: over the ten seed-level rates neither
separates from the size of the test, $p = 0.065$ and $p = 0.15$. Ten seeds do
not resolve a rate of this size, so we read the ordering and not the
significance. We are precise about what that
reproduction is worth, because it is weaker than a replication.
Table~\ref{tab:decomposition} measures the same estimand at $5.50\%$ over 100
seeds, and these ten seeds draw the same trajectories from the same builder, so
the two are not independent in the data. What differs is the surrogate draw: the
two implementations permute within the same blocks but consume randomness
differently, so on the shared ten seeds they return $9$ of $200$ and $12$ of
$200$. The agreement therefore says that the result does not depend on how the
within-block permutation is coded, which is worth knowing and is not evidence
from new data. The powered figure is the one to quote. Against injected links it is more
sensitive than conditioning by a wider margin than in the social dilemma: $46.5\%$ against
$19.0\%$ for conditioning at $c = 0.02$ and $89.0\%$ against $61.0\%$ at
$c = 0.04$, on the same window and the same pairs, with exact McNemar giving 68
against 13 and 65 against 9 ($p = 4 \times 10^{-10}$ and $p = 1 \times
10^{-11}$ on pairs, $p = 0.002$ on the ten seeds in both cases).

That separation is sharper than the social dilemma's, where the two calibrated
constructions were interchangeable, and it runs in the more useful direction: the
game with a residual after exclusion is the game where the surrogate change both
controls the error rate and, below the ceiling, finds more of the injected links
than conditioning does.

This is the evidence the recommendation needed and did not previously have. A
test that never fires would score perfectly on the null control, so the
$5.25\%$ and $5.50\%$ of Table~\ref{tab:decomposition} establish nothing on
their own; that is the standard Section~\ref{subsec:te-result} sets against the
ablation, and it applies here with equal force. The within-block surrogate
meets it in both games, and it is the only one of the three that changes
neither the data nor the estimand.

\begin{table}[htbp]
\centering
\caption{Detection rate against coupling strength (Public Goods, 10 seeds, 200
  directed pairs per cell, all four procedures on the identical pairs). The
  $c = 0$ row is the null control, where every detection is a false positive.
  The transfer entropy column is measured on the post-transient window, which is
  the window exclusion analyses, so it indexes the rows and is not a common scale
  for the two procedures that read the whole series. The final column is the
  exact McNemar $p$-value for the contrast Section~\ref{subsec:titration} rests
  the sensitivity claim on, the within-block surrogate against block
  conditioning, on the pairs where the two disagree; at the top two couplings
  they agree on every pair. The comparison with transient exclusion is quoted in
  the text and read there as a bound.}
\label{tab:titration}
\footnotesize
\setlength{\tabcolsep}{3pt}
\begin{tabular}{@{}lcccccc@{}}
\toprule
$c$ & \textbf{TE (bits)} & \textbf{Abl.} & \textbf{Excl.} &
\textbf{Cond.} & \textbf{Within blk.} & \textbf{McNemar} \\
\midrule
0 (null) & $0.00013$ & $100\%$ & $4.5\%$ & $6.5\%$ & $3.0\%$ & $0.14$ \\
0.01 & $0.00023$ & $100\%$ & $15.5\%$ & $9.5\%$ & $18.0\%$ & $0.016$ \\
0.02 & $0.00036$ & $100\%$ & $26.5\%$ & $21.0\%$ & $39.5\%$ & $2\times10^{-5}$ \\
0.04 & $0.00076$ & $100\%$ & $48.5\%$ & $58.0\%$ & $82.5\%$ & $3\times10^{-9}$ \\
0.08 & $0.00187$ & $100\%$ & $82.0\%$ & $98.0\%$ & $100\%$ & $0.13$ \\
0.16 & $0.00525$ & $100\%$ & $99.0\%$ & $100\%$ & $100\%$ & n/a \\
0.32 & $0.01442$ & $100\%$ & $100\%$ & $100\%$ & $100\%$ & n/a \\
\bottomrule
\end{tabular}
\end{table}

Two limits apply. The injected link is a copy. A learned response would differ,
and the copy fixes the functional form of the dependence while varying only its magnitude,
so a link of equal transfer entropy arising by another mechanism need not sit
at the same point on these curves. And the block count is held at 32
throughout, so the reversal at $c = 0.01$ describes conditioning as specified
in Section~\ref{subsec:conditional}. It does not describe conditioning in general.

\subsection{What Sustains the Congestion Residual}
\label{subsec:stationarity}

Everything above establishes that the transient has to be treated, and that
excluding it is insufficient in the congestion game. It does not say why. Throughout this
section we have been careful to claim only that we exclude a known transient.
We have not claimed to verify stationarity. That distinction now becomes decisive, because
the leading candidate explanation for the residual is that the congestion
series are simply still not stationary once the transient is gone.

For every action series entering the analysis we ran two complementary tests,
on the full series and on the post-transient window: the Augmented
Dickey-Fuller test, whose null hypothesis is a unit root, and the KPSS test,
whose null hypothesis is stationarity. They are used as a pair deliberately.
ADF has low power against near-unit-root alternatives, so a rejection on its
own is weak evidence, and running both allows agreement to be checked.

The 600 series per game are the learning agent's action series from each of
600 independent runs, 6 runs in each of the 100 seeds. These are the same
six-run pool the rest of the null control uses, and the rate we correlate
against below is the late-window rate of
Table~\ref{tab:windows}. Unlike the control pairs, which
are drawn from a shared pool of runs and are therefore clustered, these are
independent, so they admit a series-level test directly. We report the counts
without a multiplicity correction, since the quantity of interest is the
proportion of series in each game. The status of any one of them is not at
issue.

In the event the two tests are very unequally informative. ADF rejects the
unit root in $600$ of $600$ series on both windows of the social dilemma and
on the full congestion series, and in $562$ of $600$ on the post-transient
congestion window. It therefore separates the two games only weakly, in the
one cell where it moves at all, and KPSS is what carries the diagnosis; we
report both counts and rest the argument on KPSS. We take that as a
property of these series. The test is not defective, since binary action
sequences at an exploration floor are far from a random walk whatever else
they are. KPSS is applied to the binary series directly. Its null is level
stationarity and its statistic uses partial sums of the demeaned series,
neither assuming a continuous marginal, so what it detects here is a moving
success probability, which is the quantity of interest. Its published critical values are
asymptotic, however, and two other uses read the test absolutely. Neither reads
it comparatively: the recursive partition of Section~\ref{subsec:conditional} stops when a
segment passes, and the gate below admits a seed when every series passes. Both need a
calibrated threshold, so we simulated one. Over 2{,}000 iid binary series per cell, matched
in length to the analysed window and spanning the success probabilities the real series
show, the empirical 95th percentile of the statistic runs $0.451$ to $0.477$ against the
asymptotic $0.463$, and the test rejects $4.5$ to $5.5\%$ of series that are stationary by
construction, $5.0\%$ on average. The asymptotic values are usable here, so the absolute
readings are not an artefact of applying a continuous-data test to binary data, and the
per-series false-rejection rate the geometric argument below needs is the nominal one.

On the full series the KPSS verdict is unanimous and unsurprising. It rejects
stationarity for all 600 series in each game, confirming that the transient is
what the exclusion is there to remove. On the post-transient window the two
games separate sharply. In the Public Goods Game KPSS calls $574$ of $600$
series stationary, $95.7\%$, and in the congestion game only $341$ of $600$,
$56.8\%$. A Fisher exact test between the games gives
$p = 2 \times 10^{-62}$.

That alone would be suggestive. The stronger evidence is that the residual
tracks it seed by seed, at the level where independence is guaranteed. Taking
each seed's fraction of KPSS-stationary series against that same seed's
late-window false-positive rate, the two are negatively correlated in the
congestion game: Spearman $\rho = -0.394$, $p = 5 \times 10^{-5}$ over 100 seeds. Seeds
whose series remain non-stationary are the seeds that generate false
positives. In the Public Goods Game the correlation is absent
($\rho = -0.09$, $p = 0.36$), as it should be: there is almost no
variation left to correlate, since nearly every series is stationary and
nearly every rate is at floor.

That association has a confound we should name, because it is the same one
Section~\ref{subsec:te-result} raised about the equal-length windows. KPSS rejection and
permutation rejection are two tests sensitive to overlapping features of the same series,
and both plausibly track how concentrated the marginal is. Measured over 40 congestion
seeds, the marginal action entropy of the analysed window is in fact the stronger
univariate predictor of the residual rate, $r = +0.59$ against the stationary fraction's
$r = -0.49$, and the two predictors are themselves correlated at $r = -0.45$. Partialling
the entropy out leaves the stationarity association at $r = -0.32$, $p = 0.048$ on
$37$ degrees of freedom. It
survives, so the diagnosis is not simply the marginal in disguise, but it survives
weakened and with a second factor of at least equal standing beside it. We report the
residual as associated with two overlapping properties of the retained series. Neither one
of them explains it alone.

\emph{Separating them by construction.} Observational data from the runs cannot
do better than that, because the anneal produces the trend and the concentrated
marginal together. Synthetic series can, since the two are then set
independently, and the manipulation costs no assumption about agents: these are
Bernoulli sequences with a prescribed mean profile over the same $12{,}000$
rounds, with no learning, no game and no interaction of any kind. One property
of binary data bounds what such a test can show, and we state it and do not
work around it. For a binary variable the marginal entropy is a deterministic
function of the mean, $H = H(p)$, so a series whose mean drifts necessarily has
a drifting entropy and the two cannot be decoupled within one series. What can
be separated is the \emph{level} of the entropy from the \emph{presence of a
trend}, and that is the comparison we run.

The drifting arm takes $p(t)$ from $0.50$ to $0.021$ on the exploration
schedule's own decay, giving a pooled mean of $0.0998$ and a pooled marginal
entropy of $0.468$ bits. The matched arm holds $p(t)$ at that same $0.0998$, so
the two arms have the \emph{identical pooled marginal distribution} and
therefore identical marginal entropy, differing in the trend and in nothing
else. A third arm holds $p$ at $0.0753$, whose entropy of $0.385$ bits matches
the drifting arm's time-averaged figure, where the entropy-matched arm matches
its pooled one, and so is \emph{lower} than either. At 100 seeds and $2{,}000$ pairs, the drifting arm
flags $2{,}000$ of $2{,}000$, the entropy-matched arm $116$ of $2{,}000$
($5.80\%$, exact binomial $p = 0.090$ against the size of the test) and the
lower-entropy arm $103$ of $2{,}000$ ($5.15\%$, $p = 0.72$). The separation
between the first two exhausts double precision.

Two things follow. The entropy level is not the mechanism. Holding it fixed and
removing the trend takes the rate from $100\%$ to nominal, and lowering it
further while keeping the series stationary does nothing. And the failure needs
no agents, no learning and no game, which the introduction claims and
this measures directly. A fourth arm sharpens the mechanism: when half the
series fall and half rise, and pairs are drawn across the two groups so that
both members drift in opposite directions, the rate is again $1{,}800$ of
$1{,}800$. Agreement in the sign of the drift is therefore not what the test
responds to, which is the same conclusion the schedule
manipulation of Section~\ref{subsec:schedule} reached by desynchronizing the
anneals. What it does respond to is fixed by the next arm.

\emph{A trend, or any moving marginal?} Every profile used so far is monotone,
including the anneal itself, so ``trend'' may be narrower than the mechanism
requires: if knowing where a series sits in its own trajectory locates the
other, the marginals need only be recoverable functions of time. The
distinction decides where the remedy applies, because permuting within blocks
works precisely when the marginal is near-constant inside a block. We therefore
replaced the monotone profile with a smooth non-monotone path, a sum of
low-frequency sinusoids with random phase built in logit space so it stays
inside $(0,1)$ without clipping, drawn independently for each series and
centred on the same pooled mean. At 20 seeds and 400 pairs the ablation flags
$73.25\%$ of pairs whose true influence is zero. Monotonicity is not required,
and the honest description of the mechanism is a marginal that varies with
time. Calling it a trend claims too much.

The same construction locates the boundary of our own recommendation. Raising
the frequency until the marginal moves appreciably inside a block, from a mean
within-block range of $0.015$ under the anneal to $0.197$, takes the
within-block surrogate from $6.50\%$ to $31.75\%$, well above the size of the
test, while a constant-marginal control stays at $4.25\%$. The remedy therefore
holds when the drift is slow relative to the block and fails when it is not,
which is a property of the construction. A block
that contains a moving marginal is not exchangeable within itself. Where the drift
has a known source the block length can be chosen against it; where it does not,
this is the case the Discussion leaves open, and it is now a
measurement. The argument no longer has to carry it.

\emph{Nor is it specific to transfer entropy.} Section~\ref{subsec:schedule}
runs the same null control with plain mutual information and obtains the same
rates, so the conditioning that distinguishes TE from MI gives no protection.
What breaks is the comparison against a permutation null under a drifting
marginal; the directed statistic is where a practitioner is most likely to meet
it, and not what makes it happen.

This reframes the result reported above. The congestion residual is the guardrail
behaving exactly as the underlying theory says it should. Transfer entropy
against permutation surrogates continues to produce false positives for as
long as the series remain non-stationary, and in this game excluding the
exploration transient does not make them stationary. Something else in that
configuration sustains the violation, and Section~\ref{subsec:roster} shows
the roster carries most of what separates it from the public-goods control. We have not identified what, and
the correlation is moderate, falling well short of overwhelming, so residual
non-stationarity accounts for part but not necessarily all of the effect. What
we can now say, which we could not before, is that the diagnosis carries
practical information: run the stationarity test and read it. Where it passes,
as in the one-learner public-goods control, the guardrail suffices. Where it
fails, excluding
the transient is not enough and the analysis needs a different remedy. What
the test supplies is which of those two cases holds. It is not a pass
condition for the analysis.

\emph{What it delivers when it is applied as a gate.} A correlation across seeds is
not the same thing as applying the rule and measuring what gets through it.
Partitioning the congestion seeds by whether every series passes KPSS, those
that pass return $2$ of $60$ pairs flagged, $3.33\%$, against $234$ of
$1{,}940$, $12.06\%$, for those that do not, separating at $p = 0.021$ by a
Mann-Whitney test on the seed-level rates. The two arms sum to the $236$ of
$2{,}000$ of the late window, since the partition is over the six-run pool. In
the public-goods game the same partition separates nothing, $2.87\%$ against
$3.60\%$ at $p = 0.19$, as expected where almost every seed already passes.

Only 3 of 100 seeds clear that gate, so the passing arm rests on 60 pairs with
a Wilson interval of $[0.9\%, 11.4\%]$. The 3 is close to arithmetic. It is not
a quirk of these runs: the rule as stated passes a seed only when all six of its
series pass, and $0.568^6 = 3.4\%$, so a conjunction over series has a pass
rate falling geometrically in the number of agents. The restriction is
therefore a property of the threshold. The diagnosis is not what imposes it,
and a usable version would threshold the \emph{fraction} of series that pass.

\emph{Which fraction.} With six series per seed the gate has exactly six
settings, ``at least $k$ of 6 stationary'' for $k = 1$ to $6$. There is no
continuous parameter here, so evaluating all six is a complete enumeration. No
threshold search is involved, and we report every setting, including the ones
we do not prefer. Nothing is simulated: both inputs already exist per seed, the
fraction of its series KPSS calls stationary and its late-window rate on the
null control. The loosest setting admits every seed and reproduces the
published $236$ of $2{,}000$ exactly, which establishes that the
reconstruction is the same quantity.

\begin{table}[htbp]
\centering
\caption{Every setting of the stationarity gate in the coordination game,
  by re-analysis of stored per-seed quantities. Relaxing the conjunction to five
  of six multiplies the validating sample sevenfold and the passing seeds stay
  at the size of the test; at four of six they no longer do.}
\label{tab:gate}
\footnotesize
\setlength{\tabcolsep}{5pt}
\begin{tabular}{@{}lcrrr@{}}
\toprule
\textbf{Gate} & \textbf{Seeds} & \textbf{Passing} & \textbf{Rate} &
\textbf{Failing} \\
\midrule
$\geq 6$ of 6 & 3   & $2/60$        & $3.33\%$  & $12.06\%$ \\
$\geq 5$ of 6 & 22  & $28/440$      & $6.36\%$  & $13.33\%$ \\
$\geq 4$ of 6 & 50  & $95/1{,}000$  & $9.50\%$  & $14.10\%$ \\
$\geq 3$ of 6 & 74  & $151/1{,}480$ & $10.20\%$ & $16.35\%$ \\
$\geq 2$ of 6 & 92  & $207/1{,}840$ & $11.25\%$ & $18.12\%$ \\
$\geq 1$ of 6 & 100 & $236/2{,}000$ & $11.80\%$ & \\
\bottomrule
\end{tabular}
\end{table}

Table~\ref{tab:gate} locates the boundary. At five of six the gate admits 22
seeds and $440$ pairs, sevenfold the conjunction's sample, and the passing
seeds return $6.36\%$, which is not distinguishable from the size of the test
(exact binomial $p = 0.19$) while the failing seeds sit at $13.33\%$. At four
of six it breaks: $9.50\%$, unambiguously above the size of the test
($p = 4 \times 10^{-9}$). The gate therefore has an operating point, and we can
say where it stops working.

Both figures are pair-level, and pairs within a seed share runs, so we give the
seed-level companion the rest of the paper uses. Over the 22 passing seeds the
mean rate is $6.36\%$ with a $95\%$ $t$ interval of $[4.18\%, 8.55\%]$, which
contains the size of the test ($p = 0.20$); over the 50 seeds passing at four of
six it is $9.50\%$, interval $[7.46\%, 11.54\%]$, which excludes it
($p = 5 \times 10^{-5}$). The boundary is therefore where the pooled test puts
it. A signed-rank test on the same 22 rates rejects, but that is an artefact of
discreteness. It is not evidence: with 20 pairs per seed the attainable rates
are multiples of $5\%$, and the realized size of $4.98\%$ sits just below the
smallest non-zero one, so every seed with any detection at all outranks it.

We state the limit of that with the same precision. The Wilson interval at five
of six is $[4.4\%, 9.0\%]$, so $440$ pairs establish that the passing seeds are
not distinguishable from nominal. They do not establish equality: a true rate
of $8\%$ would be compatible with what we measured. What the enumeration settles
is the shape of the dependence, that the rate rises monotonically as the gate
is relaxed and crosses out of the nominal region between five and four of six.
In the public-goods game the gate has nothing to do: 99 of 100 seeds pass at
five of six, and the rate is below nominal at every setting, which is what one
expects where the guardrail already suffices.

\subsection{Entropy and Mutual Information on the Same System}
\label{subsec:other-measures}

\begin{figure}[htbp]
\centering
\includegraphics[width=\linewidth]{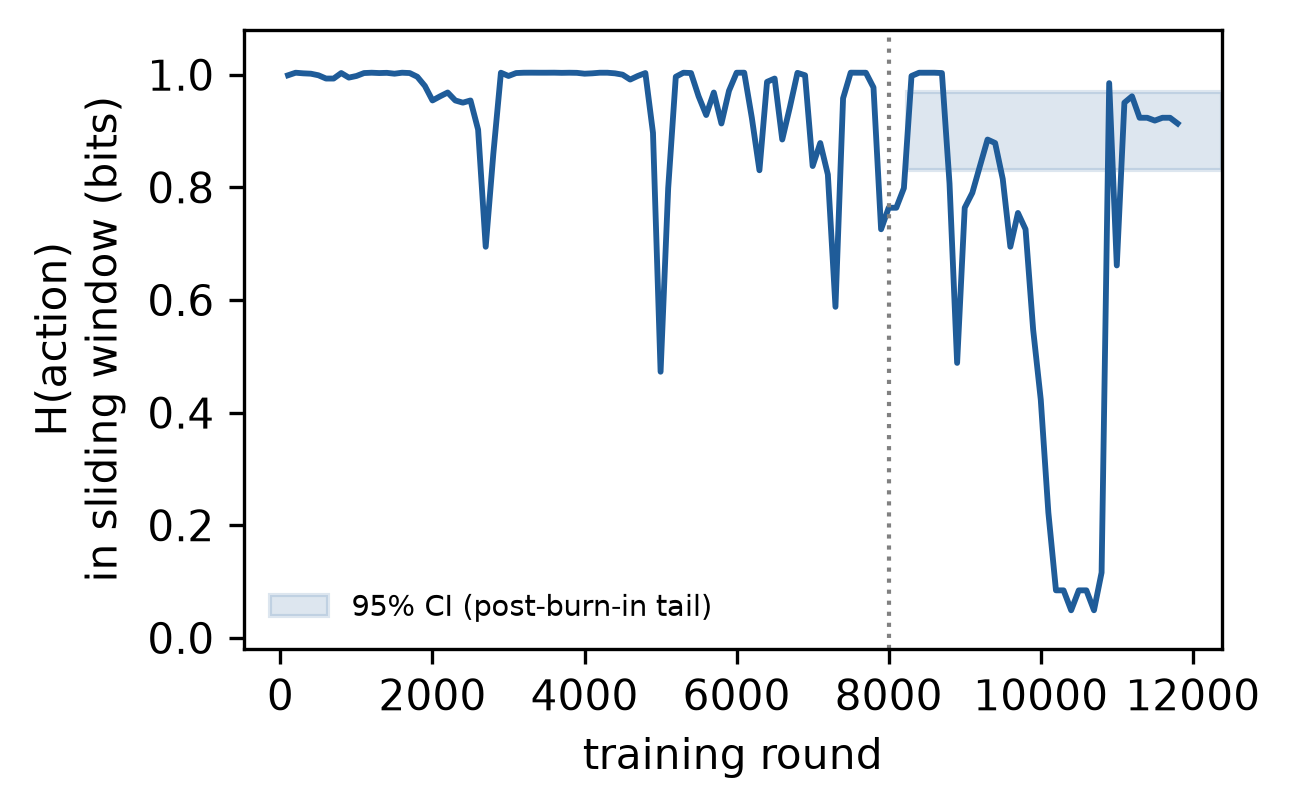}
\caption{Policy entropy $H(\mathrm{action})$ of a Q-learning agent in a
  sliding 200-round window (discrete plug-in, Miller-Madow corrected). The
  dotted line marks the burn-in cutoff; the shaded band is a 200-resample
  bootstrap 95\% CI on the converged tail. Entropy stays high and does not
  collapse monotonically, with transient near-deterministic episodes. A
  policy-entropy trace is not a stationarity test, but this pattern is
  consistent with continued co-adaptation between the two learners, that is,
  with the kind of ongoing dynamics that motivates the precondition
  Section~\ref{subsec:te-result} shows to be consequential.}
\label{fig:entropy}
\end{figure}

\paragraph{Entropy} Figure~\ref{fig:entropy} shows the sliding-window policy
entropy of one Q-learning agent. Across 100 seeds the first window measures
$1.000 \pm 0.007$~bits and the converged tail $0.866 \pm 0.239$~bits (seed-0
bootstrap 95\% CI $[0.830, 0.968]$). The value does not approach the near-zero
entropy a single-agent bandit would reach, because the two co-adapting
learners continuously perturb each other's effective environment. The figure visualizes the policy
dynamics that motivate the stationarity concern in the first place.

\begin{figure}[htbp]
\centering
\includegraphics[width=\linewidth]{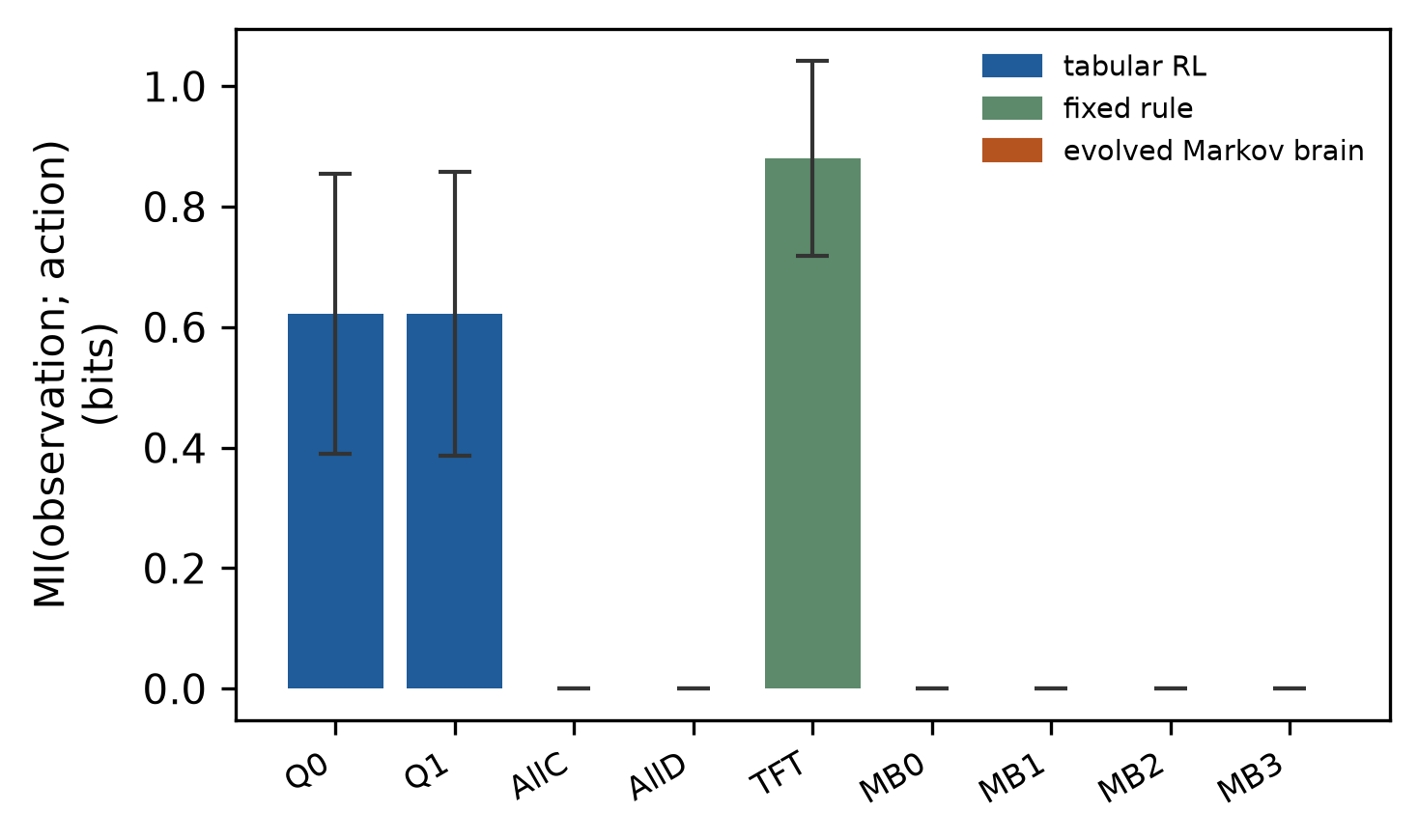}
\caption{$I(\mathrm{observation}; \mathrm{action})$ per agent (discrete
  plug-in, Miller-Madow corrected), mean $\pm$ standard deviation over 100
  seeds, coloured by architecture. The Q-learning and fixed-strategy agents
  are measured on the post-burn-in tail; the animats come from separate
  5{,}000-round evaluations of the fittest genome, measured whole, since their
  policy is fixed and has no transient to exclude. The two constant
  strategies measure exactly zero with zero variance. The learned policies vary
  by more than half a bit across seeds, and the evolved animats sit at the
  estimator's noise floor. The error bar on majority tit-for-tat understates
  its spread, which runs the full width of the measure, $0.004$ to $1.000$
  bits, because its value depends on a signal distribution the other agents
  produce.}
\label{fig:mi}
\end{figure}

\paragraph{Mutual information} Figure~\ref{fig:mi} compares
$I(\mathrm{observation}; \mathrm{action})$ across all three architectures
under the estimator recommended in Appendix~\ref{sec:mi}. The fixed strategies
are as predicted. The two constant strategies measure exactly $0$ bits with
zero variance across seeds, and majority tit-for-tat, a deterministic function
of the observation, measures $0.880 \pm 0.163$~bits: the rule is fixed, but
its input is the public signal the other agents produce, so its value moves
with them. The evolved animats sit at the estimator's floor for a reason the
raw series give directly. They converged on the free-riding equilibrium,
cooperating in $0.3\%$ of rounds, so the action series carries
$0.030 \pm 0.021$~bits of entropy over the 400 seed-and-agent combinations,
and no mutual information with the observation can exceed that.

The learned policies behave differently, and the size of the difference is the
point. The two Q-learners average $0.623 \pm 0.233$ and $0.622 \pm 0.236$~bits,
but the average is a poor summary: across 100 seeds of an identical
configuration the value ranges from $0.000$ to $0.918$~bits, essentially the
full width the measure can take here. Sixty-six of the 100 seeds encode the
public signal strongly ($> 0.6$~bits) while 16 settle on policies close to
independent of it ($< 0.3$~bits), the remaining 18 lying between. At five
seeds this looked like a clean two-regime split. At 100 it is better described
as a broad, left-skewed distribution with a substantial near-independent tail.

The reporting consequence is unchanged and, if anything, sharper: a single-run
MI value for a \emph{learned} policy characterizes that run. The algorithm is
not what it describes. A practitioner who happened to draw one of the 16 low seeds would
report a figure an order of magnitude below one drawn from the upper group, on
identical code with identical hyperparameters. The robustness-across-seeds
requirement of Appendix~\ref{sec:mi} is doing real work here.

The same effect reaches the tit-for-tat agent, whose value spans $0.004$ to
$1.000$~bits across seeds, even though that agent's decision rule is fixed. Its mutual information depends on the distribution of the public
signal, and that distribution is produced by the other agents. Even a
deterministic rule therefore has no single characteristic MI: the value is a
property of the rule together with the population it is embedded in.

One further detail calibrates the floor of the method, and it is not a noise
floor. In 19 of the 400 seed-and-agent combinations the point estimate is
marginally negative, the lowest being $-1.4 \times 10^{-4}$~bits. That value
is deterministic. Nothing stochastic produces it. Applying the Miller-Madow correction
to each of the three entropies of $I = H(X) + H(Y) - H(X,Y)$ shifts the
estimate by $(m_X + m_Y - m_{XY} - 1)/(2N\ln 2)$, where each $m$ counts
occupied cells, so on this data the shift takes one of two values: zero when
the joint table has $m_X m_Y - m_X$ occupied cells, and
$-1.443 \times 10^{-4}$~bits at $N = 5{,}000$ when one further cell is
occupied. Every negative reading we observe is that single quantum plus a
small positive plug-in value. The practical consequence is that magnitudes at
the $10^{-4}$~bit scale are not interpretable as dependence in either
direction, and should be read as zero. Significance is unaffected, because
each estimate is compared against surrogates with the same cell occupancy, so
the shift is common to the statistic and its null distribution and cancels.

\subsection{Applying Our Own Reporting Checklist}
\label{subsec:checklist}

Appendix~\ref{sec:synthesis} states six reporting requirements, which we apply
to the experiment above.
One point of alignment first, since the gap is deliberate. The estimator
discussion in Appendices~\ref{sec:mi} and~\ref{sec:te} is mostly about KSG,
variational bounds and toolboxes that do not appear in the main arms, because
those exist for continuous, high-dimensional or non-uniformly embedded data.
Here the action space is binary and the state space small, so the
Miller-Madow plug-in is the exact estimator. No compromise is involved, which
is what a test of preprocessing requires: the estimator must not itself be a
source of doubt. The continuous-action arm re-runs the diagnosis with KSG and
the failure recurs; the neural family remains untested.

\emph{(R1) Estimator or surrogate.} Every value reported here is a measurement
estimator. No variational bound (MINE, InfoNCE) is used anywhere in this
section, and none of these numbers is a training objective.
\emph{(R2) Estimator family, hyperparameters, software.} Discrete plug-in with
Miller-Madow correction throughout, applied to each of the four entropies of
the conditional-MI decomposition from its own count of occupied cells; transfer
entropy at lag 1 via that identity; 200 random-permutation surrogates per
directed pair; stationarity diagnostics with ADF at AIC-selected lag order and
KPSS with a constant regressor and automatic Newey-West bandwidth;
$\alpha = 0.05$ as the applied threshold, whose realized size at 200
surrogates is $10/201 = 0.0498$, uncorrected for multiple comparisons. That
last choice requires justification. The smallest attainable $p$-value is
$1/201 \approx 0.005$, exactly the value every significant link attains, so a
Bonferroni threshold for the 20 within-roster comparisons, $0.0025$, lies
below the floor and would need at least 400 surrogates to test at all. The six
within-roster links are therefore significant at uncorrected $\alpha = 0.05$
only. The calibration result, which is the paper's actual claim, is unaffected:
the null control measures the false-positive rate at the nominal uncorrected
level, which is the quantity being validated.
\emph{(R3) Uncertainty intervals.} Aggregate quantities are reported as
mean $\pm$ standard deviation over independently seeded replications, 100 for
the main arms and 10 or 20 for the parameter sweeps, with each table stating
its own count;
the entropy tail additionally carries a 200-resample bootstrap 95\% CI.
\emph{(R4) Surrogate significance testing.} Every transfer-entropy value is
accompanied by a surrogate $p$-value, and the null-control condition exists
precisely to check that those $p$-values are calibrated.
\emph{(R5) Preprocessing.} Actions are natively binary, so no discretization
was applied. The single preprocessing decision is the burn-in exclusion, and
Section~\ref{subsec:te-result} quantifies its effect and does not merely
declare it.
\emph{(R6) Transient treatment.} The main arms exclude it: the analysed window begins at
round $8{,}000$, chosen because $\epsilon$ reaches its floor at round $7{,}823$; the
stationarity diagnostics of Section~\ref{subsec:stationarity} report what that
window achieves in each game, including where it does not suffice. This is the
requirement the paper tests, so it is the one item of the six for which we can
say what following it buys. R6 allows four routes and the paper exercises all
four on the same data: exclusion in the main arms, the within-block surrogate
and conditioning on the driver in Section~\ref{subsec:conditional}, which
measures them separately and together, and the ensemble estimator in
Section~\ref{subsec:ensemble}.

\subsection{Reproducibility and Independent Verification}
\label{subsec:repro}

All figures and numbers in this section are produced by a self-contained
analysis script, released as supplementary material together with the seed
sweep and the complete JSON output. Two stored outputs are exceptions and we
name them. \texttt{all\_experiments\_summary.json} and
\texttt{cutoff\_sensitivity\_summary.json} were written before the run
indexing of the eight-run null pool was corrected, so the released code
rebuilds that pool differently and will not reproduce their null-control
cells. The rates the manuscript took from the first were withdrawn for a
related reason, set out in Section~\ref{subsec:design}; the second enters only
as the paired cutoff contrast of Section~\ref{subsec:second-game}, whose two
arms shared one pool, and as the summary of that contrast in the introduction
and the conclusion, which name the pool where they state it. Which pool an arm
drew on is checkable from the released code rather than from its pair count:
\texttt{run\_cutoff\_sensitivity.py} calls \texttt{te\_validation} with that
function's default of eight control runs, and the 30 pairs it reports per seed
are a draw from the 56 that eight runs admit, not the exhaustive set of a
smaller pool. The real-roster cells of the first are unaffected, since
that roster is built on a different index. Every figure the manuscript reports
as a calibration result comes from the six-run pool and rebuilds from the
released code. To guard against implementation error, the
headline mutual-information values and a spot check of transfer-entropy pairs
were re-derived by a second, separately written script that shares no code
with the first or with GameBrains' metrics module, using a different counting
implementation and joint-symbol encoding. All re-derived values matched to
floating-point precision. The same pairs were then cross-validated against
\texttt{pyinform}, the Python interface to the independently authored Inform
library~\cite{moore2018inform}. Three of five pairs matched to floating-point precision
and the remaining two agreed to within $3 \times 10^{-4}$~bits, a gap
consistent with that toolkit not applying the same finite-sample bias
correction. An estimation error in either implementation would not produce it.
On the scale of the null control, whose values average
$1.4 \times 10^{-4}$~bits, that gap exceeds the entire range in which false
positives live. It propagates to no conclusion here because every decision is
made against surrogates from the same implementation. No absolute value
enters, which is itself a reason not to compare magnitudes at this
scale across toolkits.

For exact reproduction, the software stack is as follows. All results were
produced with Python 3.11.9, NumPy 2.4.6, SciPy 1.17.1 and statsmodels 0.14.6
on Windows 11 (x86-64). No GPU is used. Every estimator is deterministic
given its seed, and none of them dispatches work across threads, so results
do not depend on thread count; the two are separate properties and we rely
on the second. The environment and
agent implementations come from GameBrains~\cite{gamebrains2026} at commit
\texttt{b01cdd7}, version 0.0.1, and the integrated-information values use the
copy of PyPhi vendored with that release. Scripts must be run against the
GameBrains virtual environment. A bare interpreter will not do, since the
vendored dependencies are resolved from it. Randomness enters at three points,
seeded explicitly and independently: environment dynamics, agent policy
initialization and action sampling, and surrogate generation. All three use
NumPy's \texttt{default\_rng} with integer seeds derived arithmetically from
the replication index and the stream or pair index, so re-running a script
reproduces its JSON output bit for bit. The independence the null control
relies on is between runs, and the run seeds are 10 apart within a replication
and 100 apart across replications, each one hashed through a seed sequence, so
no two PCG64 states could overlap. Training runs for a fixed 12{,}000 rounds with no early stopping and no
checkpoint selection, so nothing is selected between training and analysis,
and hyperparameters are set in the released scripts and not a separate
configuration file.

\section{Discussion}
\label{sec:discussion}

\paragraph{Scope of what was validated} One guardrail was tested. The
framework was not. The transfer-entropy stationarity precondition now rests on a
measured comparison against ground truth. The remaining recommendations in
Appendix~\ref{sec:synthesis} do not, and should be read as what they are:
distilled practice. None of them is a validated instrument. The
reporting-guideline literature outside AI/ML has made the same distinction
explicit, warning against treating a checklist's adoption as evidence for the
quality of the decisions it informs~\cite{puljak2019,logullo2020}. We make no
claim that using the selection table or the reporting checklist measurably improves
selection accuracy or inter-practitioner agreement. Establishing that would
require a controlled user study.

\paragraph{Limitations of the experiment} The boundaries of the result should
be stated plainly, and they are a condition. They are not a domain. What produces
the failure is a marginal that varies with time in both series over the window
analysed, which the synthetic arm demonstrates with no agents and no game. The
claim therefore covers any system analysed while that holds, and we make none
about multi-agent reinforcement learning in general. Runs analysed after
convergence fall outside it, as do architectures beyond those measured here:
tabular Q-learning, DQN and a continuous-action REINFORCE agent for the
transfer-entropy result, and tabular, fixed-rule and evolved-network agents
for the mutual-information comparison. Agents whose exploration rate is fixed
are not outside it. Section~\ref{subsec:schedule} measures them, the rate
falls to $23.5\%$ and stays above the size of the test, and what we cannot
say is what sustains that remainder. The two games are evidence
that the effect is not specific to a payoff structure. They are not evidence
that it is universal.
The roster inside each run matters at least as much as the game, and we found
that only because a reviewer asked. Our two games were not a two-point sample
of payoff structures; they were a two-point sample of payoff structure crossed
with roster, and the roster carried most of the difference
(Section~\ref{subsec:roster}). The same question applies to every other factor
we held fixed without varying it.
Both environments are repeated games with binary actions
and small discrete observation spaces, chosen because they make ground truth
available. Neither is claimed to be representative. A DQN arm shows the
effect is not confined to tabular learning, and a continuous-action arm with a
kNN estimator shows it is not confined to discrete data either. Those
arms settle less about the remedy. In the continuous setting neither transient
exclusion nor block conditioning reaches the size of the test, and the
within-block surrogate is the only one of the three that does, at $10.0\%$ on 40
pairs; but 40 pairs cannot separate it from the others, so what transfers is
that the recommendation is not excluded there, which is weaker than winning there.
With $1{,}500$
post-transient samples we also cannot separate surviving drift from the variance
of a neighbour-based estimator near its sample limit. Two games establish that
the effect is not specific to one payoff structure but do not establish
universality, and richer observation channels remain untested. Everything here
is at lag 1. Multivariate conditioning is no longer untested and does not
rescue the bivariate result: conditioning the statistic on one or two other
co-drifting runs leaves the null-control rate at $100\%$ on all $2{,}000$ pairs
of each arm (Section~\ref{subsec:conditional}), so the failure is not an
artefact of analysing pairs in isolation. What remains untested is non-uniform
embedding, conditioning across more than one lag, and conditioning sets larger
than two. The schedule
manipulation isolates that mechanism and bounds it: desynchronizing the
schedules leaves the rate unchanged, so synchrony is not the operative factor.
It does not settle the residual under constant exploration, larger for DQN
than for tabular agents, which we attribute to learning-induced drift without
isolating its source.
The equal-length comparison in Section~\ref{subsec:te-result} controls for
sample size in both games and localizes the congestion residual to something
other than the exploration transient, and the ADF and KPSS diagnostics in
Section~\ref{subsec:stationarity} identify residual non-stationarity as what
sustains it. That diagnosis is correlational. Nothing was manipulated to establish it: we
show that the seeds whose series stay non-stationary are the seeds that produce
false positives ($\rho = -0.394$), but the correlation is moderate, so
residual non-stationarity plausibly accounts for part of the
effect, and we do not identify what in the coordination dynamics produces it.

What follows from that diagnosis is a diagnostic. It is not a gate. Requiring
every one of a seed's six series to pass admits 3 seeds of 100, which is
arithmetic more than information. Relaxing it to five of six admits 22 and the
passing seeds stay at the size of the test, while at four of six they no longer
do. The stationarity test therefore locates how much drift a
construction is being asked to absorb, and that is how we use it. It is not a
pass condition for the analysis.
Finally, a null control establishes calibration. Power is a separate question,
and Section~\ref{subsec:titration} measures the difference. Varying the strength
of an injected link shows that detection by exclusion falls below one half at
roughly $7.6 \times 10^{-4}$ bits, where conditioning still returns $58.0\%$
and the within-block surrogate $82.5\%$, and that all three fall under a half
by about $3.6 \times 10^{-4}$ bits, so genuine influences weaker than that are
not reliably recoverable at 12{,}000 rounds. The construction we recommend is
therefore a guard against false positives. It guarantees nothing about
sensitivity, and a null result under it should be reported as such. It is not
evidence of independence. The same experiment answers a question the calibration
results could not: the ablation reports $100\%$ whether or not a link exists,
which is what disqualifies it, and the within-block surrogate is measurably more
sensitive than conditioning in the mid range, on the same pairs over the same
window.

\paragraph{When not to use IT measures} TE and MI are on their own
insufficient evidence for interventional causal claims: both are
observational, both can be non-zero under purely correlational processes, and
surrogate testing does not convert either into causal identification. The six
within-roster links of Section~\ref{subsec:te-result} are a concrete instance,
every one real as a predictive relation and none a direct causal link, since
no agent observes another's action. The remaining anti-patterns are named in
Appendix~\ref{sec:background} and recur often enough to repeat: kNN estimators on
high-dimensional data without variance assessment, MINE or InfoNCE reported as
measurements~\cite{tschannen2020}, and KL divergence treated as a distance.

\paragraph{Open problems} Reliable estimation of mutual information in high
dimensions remains unsolved: neural estimators are biased and
variance-prone, parametric models are misspecified, and no agreed benchmark
exists for continuous high-dimensional MI. Transfer entropy under
non-stationary and non-linear conditions with latent confounders is an active
area without consensus methods, and the failure mode quantified here is one
instance of a broader problem. Section~\ref{subsec:surrogate} rules out the
simplest fix, since neither of the standard surrogate constructions survives a
shared trend. Section~\ref{subsec:detrend} rules out the next-simplest, since
re-thresholding a detrended binary series is the identity map, and shows that
the one transform that does work, first differencing, answers a different
question. What would be needed instead is a null model that reproduces the
trend while destroying only the dependence between series. That object is not
missing from the literature so much as absent from multi-agent practice.
Lucio et al.~\cite{lucio2012} construct surrogates that preserve trends within
the Fourier-based family, which we did not test. The ensemble estimators developed for
neuroscience~\cite{wollstadt2014} sidestep the problem differently, estimating across
repeated realizations where the pooled statistic runs along one series, and independent
training runs are the natural multi-agent analogue of experimental trials.
Section~\ref{subsec:ensemble} runs that construction. It returns rates at the size of the
test in both games on the untrimmed series, but on 30 tests, whose intervals are too wide to
claim an improvement on the remedies that need no ensemble. It shows that a statistic which
never pools over time is not defeated by the drift that defeats pooling; it does not show
that this is the best way to handle the case where the driver is unknown.

What we did test, in Section~\ref{subsec:conditional}, is the option those two
families do not cover: conditioning on the driver, where they model it.
Where the drift comes from a schedule the experimenter set, the driver is
observed, with nothing latent to infer, and blocking on training time brings
the rate back into the nominal region on the full series without discarding
data. That holds in the one-learner social dilemma. In the coordination game,
which has four learners, it does not,
and the same conditioning reaches $8.55\%$ and stops: it removes the drift its conditioning
set describes, and there a substantial part of the drift is not the schedule.
Changing the surrogate alone does better on both counts. It reaches the size of
the test in both games, and the titration in
Section~\ref{subsec:titration} finds it more sensitive than conditioning in the
mid range, and bounds above what exclusion recovers there, which is the reverse
of what their cost would
suggest: leaving the data, the statistic and the estimand alone costs no
sensitivity against the remedy that changes more of them.
The open problem is therefore narrower again. It is not the case where the trend has no
known source, since the ensemble construction of Section~\ref{subsec:ensemble} covers that
in principle, at a cost in runs that is why the conclusion stops short of recommending it. What
remains open is the single-run case: where only one trajectory exists, no ensemble can be
formed, and the drift has a source the experimenter cannot write down. Every remedy we
have measured needs either the driver or the ensemble.

The design's main calibration evidence draws its pairs from separate runs, which
is what makes their ground truth unimpeachable and is not the situation a
practitioner is in. Section~\ref{subsec:withinrun} closes that gap and does not
concede it: an agent that anneals on the training schedule while discarding
every observation it is handed drifts like the others and cannot be influenced
by any of them, so a single match contains directed pairs whose true answer is
zero and whose marginals are not degenerate. The failure reproduces there at
full force, $100\%$ and $93.83\%$, and both remedies control it, which is the
result that transfers to how the measure is actually used. What that arm still
does not cover is the case of a single run with no such agent in it and a drift
whose source the experimenter cannot write down. That case is now measured. We report the failure
instead of conceding it as an untested limit, and it is where the
recommendation fails hardest:
Section~\ref{subsec:stationarity} drives the marginal on a smooth non-monotone path
fast enough to move inside a block, and the within-block surrogate rises from
$6.50\%$ to $31.75\%$. The diagnosis holds there, since the ablation fails too,
but the remedy needs a block over which the marginal is near-constant, and
choosing one needs some knowledge of the timescale on which the drift moves.
Where that knowledge is absent, none of our remedies has a handle. Integrating IT measures with causal graphical
models~\cite{pearl2009} may narrow the gap between observational and
interventional analysis.

\section{Conclusion}
\label{sec:conclusion}

We set out to test whether one information-theoretic reporting guardrail earns
its keep. Omitting one preprocessing step, the exclusion of the
non-stationary training transient, raises the transfer-entropy false-positive
rate to $100.00\%$ and $99.95\%$ in the two games examined. Essentially every
null-control pair in every seed is incorrectly flagged as exhibiting directed
influence, between agents that never shared an environment. In the coordination game the ablated
test additionally flags $99.8\%$ of genuine pairs, so it barely discriminates
at all. Because the social dilemma's real roster yields the same six links
under both procedures, the precondition sacrifices almost no sensitivity there.

A direct manipulation identifies the mechanism. Pinning exploration at a
constant rate cuts the full-series rate from $100\%$ to $23.5\%$, while
staggering the schedules or giving each agent its own decay rate leaves it at
$100\%$: what matters is that both marginals move over the run. Schedule
synchrony plays no part.
Replacing the tabular learners with DQN agents changes neither the failure nor
the remedy, $100\%$ and then $1.5\%$, though the neural residual under
constant exploration is larger, $44.0\%$ against $23.5\%$. Practitioners
applying transfer entropy to populations of independently trained agents must
treat the known training-driven non-stationarity before pooling.
The rule at the end of this section says which treatment we would reach for.

The remedy is less effective in the coordination game. Excluding the
transient brought the rate to $3.05\%$ in the social dilemma, which is
indistinguishable from the estimator's own floor at that marginal. The
procedure adds no conservatism of its own. Nor is the cutoff delicate: on the earlier
eight-run pool, where the two cutoffs are paired on the same trajectories, an error of $800$
rounds into the tail of the anneal leaves that pool's rate statistically unchanged. In the coordination game it stalled at
$11.8\%$, and retreating further from the transient made matters slightly
worse. It did not help.

Formal stationarity diagnostics go with that difference. After exclusion KPSS
calls $95.7\%$ of the social-dilemma series stationary against only $56.8\%$
of the coordination series, and within a game the fraction still
non-stationary predicts the residual rate across seeds. Across arms it does
not: the four-learner social dilemma is the least stationary of the three and
has the lower rate. Exclusion leaves those series non-stationary, so it does
not satisfy the stationarity requirement there, which is a statement about
what the precondition delivers. It says nothing about the estimator. What sustains
the violation we have not identified, and the correlation is moderate enough
that it may account for only part of the residual.

Changing the null model is not futile, but the two obvious changes are. Substituting a
circular time-shift surrogate, which preserves each series' own temporal structure, made
almost no difference: $100\%$ and $98.2\%$ against $100\%$ and $99.9\%$ for
permutation. Both break the alignment between series across the whole run, where the drift
is steep. A third does not: permuting the source within blocks of training time breaks the
same alignment at a resolution where the drift is flat, and reaches $5.25\%$ and
$5.50\%$, the size of the test in both games, changing nothing about the statistic or
the data. It is the simplest remedy we measured. In those two games that is control
and not merely reduction: an exact binomial test against the realized size does not
place either rate above it. It is not control everywhere, and the distinction matters
more than the headline. On a four-learner public-goods roster the same construction
reduces the rate to $9.80\%$ and no further, which the size of the test excludes, and
there none of the three remedies reaches that size. Reducing a rate and controlling it
are separate claims, and only the first survives every configuration we ran.

Conditioning is a different route to the same end. The drift here is an
observed confounder, since the schedule is set by the experimenter. Nothing
about it is latent, and an observed common driver can be conditioned on where a
latent one has to be removed. Transfer entropy conditional on a coarse index of training time, with
surrogates permuted within each block, brings the full-series rate from
$100.00\%$ to $5.05\%$ while recovering all 60 of the 60 connected pairs, and the
block count need not be chosen by hand: splitting recursively until KPSS
accepts each segment selects a median of 13 blocks and reaches $6.5\%$. At
$2{,}000$ pairs that partition is calibrated, $5.45\%$ against the fixed count's
$5.15\%$ on identical pairs, but it detects fewer weak links, so it removes a
parameter, and does not improve on one. That
result is confined to the one-learner social dilemma. Run on the coordination
game, the
same conditioning brings the full series from $99.95\%$ to $8.55\%$ and no
further, and moves the post-transient residual only from $11.5\%$ to $9.5\%$,
both still above nominal. Conditioning removes the drift its conditioning set
describes, and in that game a substantial part of the drift does not come from
the schedule.
Weighting blocks by bootstrapped precision instead does not improve on it,
$9.0\%$ against $6.5\%$ on the same 200 pairs, which does not separate the two.
Differencing controls the rate just as well, at $5.0\%$, but estimates
transfer entropy between the differenced series, a different quantity with its
own null hypothesis. A magnitude cutoff, the cheapest alternative of all, does
not substitute either: the false positives sit at $0.012$~bits, overlapping
the genuine links, and removing them all costs $21.7\%$ of those links against
the $7$ in $600$ the precondition costs. Where the source of the drift is
unknown, the estimator can avoid needing it. Estimating across independent runs at matched
rounds, where the pooled statistic runs along one, never pools over time and so
never assumes stationarity~\cite{wollstadt2014}: on the full untrimmed series it returns $3.3\%$ and
$6.7\%$ where the pooled statistic returns $100\%$ and $99.3\%$. That is a
demonstration of principle. We do not recommend it, since an ensemble buys one test
from many runs and 30 tests cannot separate it from the cheaper remedies. Trend-preserving
surrogates~\cite{lucio2012} remain untested.

Calibration cannot settle which remedy to prefer, so we varied the strength of
a genuine link. The ablation reports $100\%$ whether the link is absent or
strong, so its apparent sensitivity carries no information. All three corrected procedures lose weak
links, exclusion first at about $7.6 \times 10^{-4}$ bits and the other two by
about $3.6 \times 10^{-4}$, so a null result under this guardrail is not
evidence of independence. The within-block surrogate is the more sensitive of
the two constructions that share a window, detecting 60 links that conditioning
misses against 11 the other way at $c = 0.04$, on identical pairs and at rates
that do not separate on the null control. It finds 83 that exclusion misses
against 15, which bounds that gap rather than measuring it, since exclusion
analyses a shorter window carrying less of the link. That
holds against an injected lag-1 link. Where the influence instead travels on a
slow ramp, as in the within-run control, the within-block null absorbs it and
exclusion is the more sensitive of the two.

Two further results concern how single-run values should be read. Mutual
information between observation and action ranged over the full width of the
measure, $0.000$ to $0.918$~bits, across 100 seeds of one learned-policy
configuration, so a single-run value characterizes the run alone. Nothing in it
describes the algorithm. Even the fixed conditional rule varied nearly as widely, because
its value depends on a signal distribution the other agents produce. The
evolved animats instead returned mutual information indistinguishable from
zero, which the recommended diagnostic explains at once: they converged on the
free-riding equilibrium and emit an almost constant action, whose
$0.030$~bits of entropy caps the measure by construction, where the same rule
in the same game averages $0.880$~bits. A near-zero reading is a statement
about the policy, provided the marginal entropies
are reported beside it.

Around this experiment we have provided a condensed account of four
foundational measures, a selection table, and a
minimum reporting checklist applied to our own results item by item. Measures outside this scope, Kolmogorov complexity and Minimum
Description Length, directed information, and partial information
decomposition, are excluded for reasons of estimator maturity or interpretive
consensus discussed in Section~\ref{sec:intro}.

Three extensions follow, the first two as separate studies. Neither is a gap
that could have been closed here. Whether the failure and the remedy carry
over to centralized-training algorithms such as QMIX, MAPPO or MADDPG is open,
and it is a substantial question: the centralized
critic is a second candidate source of shared drift, standard benchmarks do
not supply ground truth by construction, and the DQN arm here already costs
two orders of magnitude more compute per seed than the tabular one. The
analysis is at lag 1 throughout. Conditioning on other agents' series does not
rescue the bivariate result, since it leaves the null-control rate at $100\%$,
but non-uniform embedding and conditioning across several lags, for which mature
tooling exists~\cite{wollstadt2019}, would test whether the remedy scales to
indirect influence among more than two agents. The third extension is immediate. The remaining
guardrails in this paper are each currently an assertion, the same
$2 \times 2$ design applies unchanged to every one of them, and each is
checkable.

The rule that follows from all of this is short enough to apply without a
checklist, and it is not the rule we set out to write. Do not begin by cutting
the data. Permute the source within blocks of training time, which changes the
null model and leaves the series, the statistic and the estimand alone; it is
the only one of the three at or below the size of the test in both
games. Check that it is; do not assume it. The configuration that defeats it, four independent learners in
the social dilemma, defeats all three. Exclusion is the lowest rate there, and
the paired test does not separate it from ours. Against an injected lag-1 link
the within-block surrogate is nonetheless the more sensitive of the two that
share a window, and it bounds above what exclusion recovers from the shorter
one. It has one
condition and we measured it: the
block must be long enough to matter and short enough that the marginal is
near-constant inside it. Driving the marginal on a path that moves appreciably
within a block takes the same construction from $6.50\%$ to $31.75\%$, so where
the drift is fast relative to any usable block, this remedy fails with the
others. Then test the remaining series for
stationarity, and treat the result as a diagnostic. It is not a gate. It tells
you how much drift the construction is being asked to absorb, and within a
game the seed-level rate tracks that fraction. Across arms it does not, so
read it as a measure of the load and never as a ranking. Excluding the transient remains correct and is the
simpler thing to explain, but it discards two thirds of every run, it does not
reach the size of the test in the coordination game, and its apparent margin
below nominal in the social dilemma is the estimator's floor. The procedure
contributes nothing to it. Conditioning on training time is worth its
extra machinery only where the schedule is what the series are drifting with.
A null result under a failed stationarity test should still be
read as inconclusive. It is not evidence of independence.

\appendices
\section{Background: Notation and the Four Measures}
\label{sec:background}

\paragraph{Notation}
We use uppercase $X, Y, Z$ for random variables, lowercase $x, y, z$ for
their realizations, and $p(x)$ for probability mass or density functions
(context disambiguates). Entropy: $H(X) = -\mathbb{E}[\log p(X)]$. KL
divergence: $D_{\mathrm{KL}}(p \| q) = \mathbb{E}_p[\log(p/q)]$.
Mutual information: $I(X;Y) = H(X) - H(X|Y) = D_{\mathrm{KL}}(p_{XY}\|p_Xp_Y)$.
Transfer entropy: $T_{Y\to X} = I(X_{t+1}; Y_t^{(k)} | X_t^{(l)})$.
All quantities are non-negative except differential entropy, which can be
negative. Logarithms are base 2 and all reported values are in bits.

\paragraph{Two roles an IT quantity can play}
Throughout this paper we distinguish two roles that IT quantities can play
in an algorithm. An \emph{estimator} (measurement tool) approximates the true
quantity from data and is used for analysis, evaluation, or scientific inference.
A \emph{bound/objective} (training surrogate) is a variational approximation
used as a loss function; it may or may not be a good measurement of the true
quantity. MINE~\cite{belghazi2018} and InfoNCE~\cite{oord2018} supply variational
lower-bound estimates that are widely used as training objectives; their
numerical values are not automatically calibrated point estimates of the
mutual information, and the bound can be arbitrarily loose depending on the
variational family. This distinction
must be stated explicitly in any paper using IT quantities.

\paragraph{Data regime factors}
Four properties of the data drive the choice of estimator.

\emph{Discrete versus continuous.} Plugin (empirical-frequency) estimators are
consistent for discrete data but carry a negative finite-sample bias,
correctable via the Miller-Madow adjustment. For low- to
moderate-dimensional continuous data, kNN-based estimators are a common
nonparametric choice.

\emph{Dimensionality.} Estimator variance grows exponentially with $d$, and
kNN methods become unreliable for MI well before $d = 20$, and the practical ceiling reported in applied work is often nearer $d = 10$.

\emph{Time series versus iid.} TE requires a temporal embedding. Stationarity
cannot be silently assumed, and known non-stationary transients should be
identified and treated before the estimate pools them, either by excluding
them or by a null model that absorbs them.
Section~\ref{sec:experiment} examines this in detail.

\emph{Observational limits.} TE and MI are observational statistics. Neither
establishes interventional causality without additional assumptions.

\paragraph{The four measures} The subsections that follow take each measure in
turn: entropy, KL divergence and cross-entropy, mutual information, and
transfer entropy. Each is treated through the three questions the experiment
depends on, namely what the measure answers, which estimator suits the data
regime, and what its most consequential misuse is. Transfer entropy is last
because it is the one the experiment puts under test, and the stationarity
requirement introduced there is the guardrail Section~\ref{sec:experiment}
evaluates against ground truth.

\subsection{Entropy}
\label{sec:entropy}

\paragraph{Definition and role}
Shannon entropy $H(X) = -\sum_x p(x)\log p(x)$ quantifies the average
uncertainty (or information content) of a discrete random variable $X$. The
continuous analogue, differential entropy, replaces the sum by an integral and
can take negative values~\cite{shannon1948}. In AI practice entropy answers
the question \emph{how uncertain is this distribution?}, which recurs in
decision-tree construction (information gain maximizes entropy
reduction)~\cite{quinlan1986}, exploration-exploitation tradeoffs
(maximum-entropy policies~\cite{haarnoja2018}),
uncertainty quantification for generative models, and label-smoothing
regularization. Differential entropy additionally appears in density estimation
and as the free term in mutual information decompositions.

\paragraph{Estimator choice}
For univariate ($d=1$) continuous data, spacing-based estimators
(e.g., the Vasicek estimator) achieve lower bias than kNN and KDE-based
alternatives~\cite{madukaife2024}. The construction does not extend
to $d \geq 2$, where kNN (Kozachenko-Leonenko) and KDE are the main options.
Evidence on which dominates is mixed: Madukaife and Phuc \cite{madukaife2024} find kNN
generally inferior to KDE for $d = 2, 3, 5$, while Alvarez-Chaves et al. \cite{alvarezchaves2024}
find kNN tends to outperform KDE overall, especially with sufficient data.
kNN is often the more convenient default because it is defined for all $d$,
requires no bandwidth selection, and has bias that decays predictably as a
function of $k$ and $N$. Which estimator is more accurate remains
case-dependent. Uniformization-based approaches offer additional
bias reduction by transforming data to uniform marginals before estimation~\cite{ao2023}. Binning (plugin) estimators are simple
but introduce discretization artifacts whose magnitude depends
sensitively on the bin-width rule. They are not recommended without explicit
bias correction for small or high-dimensional samples.

\paragraph{Failure modes and guardrails}
Finite-sample bias is the dominant concern. For discrete data, the plugin
estimator is negatively biased at finite $N$ (the Miller-Madow correction
partially removes this bias). For differential entropy estimators, bias
direction is not uniformly negative: it depends on the density and estimator
family, and can be positive or negative depending on local curvature. In both
cases, bias grows with dimensionality. The standard reporting protocol has three parts. State whether $X$ is discrete
or continuous and how it was discretized or normalized. Specify the estimator
family and its hyperparameters, meaning the bin rule or the value of $k$ for
kNN. Provide bootstrap or subsampling uncertainty intervals together with
sensitivity to preprocessing choices. As dimensionality grows relative to
sample size, finite-sample bias and estimator variance dominate, so dimension
reduction before estimation is often preferable to direct high-$d$
estimation. No single cutoff separates the two regimes.

\subsection{KL Divergence and Cross-Entropy}
\label{sec:kl}

\paragraph{Definitions and relationship}
The Kullback-Leibler (KL) divergence $D_{\mathrm{KL}}(p \| q) =
\sum_x p(x)\log\frac{p(x)}{q(x)}$ measures the information lost when
distribution $q$ is used to approximate $p$~\cite{kullback1951}. It is
non-symmetric and non-negative, equalling zero only when $p = q$.
Cross-entropy $H(p,q) = -\sum_x p(x)\log q(x) = H(p) + D_{\mathrm{KL}}(p\|q)$
decomposes into the true entropy of $p$ plus the KL divergence. In
classification, minimizing cross-entropy with respect to model parameters $q$
is equivalent to minimizing $D_{\mathrm{KL}}(p_{\mathrm{data}}\|q_\theta)$
when the true label distribution $p$ is fixed. KL divergence also appears as the
regularization term in variational autoencoders (VAEs), constraining the
learned posterior toward a prior~\cite{kingma2014}, and as a policy-update
constraint in trust-region reinforcement learning. TRPO imposes a hard KL
bound between consecutive policies~\cite{schulman2015trpo}, while PPO replaces
this with probability-ratio clipping as its primary mechanism but
optionally adds a KL penalty, and KL regularization more generally improves
the optimization landscape of RL objectives~\cite{schulman2017ppo,lazic2021}.

\paragraph{Estimator choice}
For discrete distributions with shared support, the plugin estimator is
straightforward to compute from empirical frequencies, but it is exact only
for those empirical distributions: as an estimate of the population
divergence it carries finite-sample bias, for the same reason the plugin
entropy estimator does. For continuous distributions, density-ratio estimation
is standard: train a classifier to distinguish samples from $p$ and $q$, then
read off the KL from the classifier's log-odds. Discriminators in reproducing
kernel Hilbert spaces (RKHS) offer lower variance and improved numerical
stability compared to unconstrained neural discriminators, particularly when
the two distributions overlap poorly~\cite{ghimire2021}. Normalizing-flow
approaches that optimize the reverse KL require careful step-size choice to
avoid gradient path issues~\cite{vaitl2022}. When $p$ and $q$ have disjoint
or near-disjoint support (common in generative modeling), KL is undefined
or infinite. Jensen-Shannon divergence or Wasserstein distance are more
appropriate in those regimes.

\paragraph{Failure modes and guardrails}
Asymmetry is the primary conceptual hazard: $D_{\mathrm{KL}}(p\|q)$ penalizes
regions where $p > 0$ but $q \approx 0$ (forward KL forces coverage), while
$D_{\mathrm{KL}}(q\|p)$ penalizes regions where $q > 0$ but $p \approx 0$
(reverse KL encourages mode-seeking). Misidentifying which direction is in
use leads to qualitatively wrong conclusions. Numerical instability from
near-zero probabilities must be addressed with additive smoothing or lower-clip
on $q$. Calibration checks (whether the reported divergence is from a
measurement estimator or a training bound) are essential for reproducibility.

\subsection{Mutual Information}
\label{sec:mi}

\paragraph{Definition and role}
Mutual information $I(X;Y) = H(X) - H(X|Y) = D_{\mathrm{KL}}(p(x,y)\|p(x)p(y))$
quantifies how much knowing $Y$ reduces uncertainty about $X$, symmetrically.
It captures non-linear dependence that correlation misses, making it the natural
tool for feature selection (does feature $X$ carry information about label $Y$?),
unsupervised representation learning via MI maximization between local and global
features~\cite{hjelm2019}, and the Information Bottleneck tradeoff between
compression and prediction~\cite{tishby2000,tishby2015}. Conditional MI $I(X;Y|Z)$ extends
this to controlled comparisons. In cognitive neuroscience, MI between a stimulus
and a neural spike train quantifies coding efficiency, meaning the fraction of
the response entropy that is stimulus-driven~\cite{borst1999}.

\paragraph{Estimator choice}
The Kraskov-St\"ogbauer-Grassberger (KSG) estimator~\cite{kraskov2004} is the
community standard for low-to-moderate-dimensional continuous data (practical
rule of thumb: $d \lesssim 20$). It uses $k$-nearest-neighbour statistics and
achieves low bias under mild density conditions. Practitioners must report $k$
and verify robustness. For high-dimensional data, neural estimators such as
MINE~\cite{belghazi2018} (Donsker-Varadhan lower bound) and
InfoNCE~\cite{oord2018} (contrastive lower bound) scale efficiently, but
they are \emph{bounds/objectives}. A measurement estimator is a different
instrument. These are appropriate as training signals but should not be used as unbiased MI
measurements without validation~\cite{tschannen2020}. The direction of the
inference matters here. Because both are lower bounds, a low value does not
establish low mutual information, and InfoNCE in particular cannot exceed
$\log N$ for batch size $N$, which caps it well below the true value in
exactly the regime where representations are informative. For interpretability
analysis in latent spaces, GMM-MI~\cite{piras2022} provides a parametric
estimator robust to discrete and continuous mixtures. When $Y$ is categorical
and $X$ is continuous, decision-forest estimators~\cite{perry2019}
outperform KSG in high-dimensional and mixed-scale settings by replacing
fixed-radius kNN with adaptive partitions.

\paragraph{Failure modes and guardrails}
The curse of dimensionality is the primary hazard: KSG variance grows
exponentially with $d$, and neural estimators may converge to vacuously loose
bounds. Leakage (data from the test set influencing the estimate) artificially
inflates MI in self-supervised evaluation. A third hazard is easy to overlook
because MI carries no explicit time argument. Estimating MI by pooling paired
samples drawn across a period in which both marginals drift will report
dependence that is an artefact of the pooling, since knowing where a sample
sits in one trajectory locates it in the other.
Section~\ref{subsec:schedule} measures this for mutual information itself, on
data where the true value is known to be zero. Whenever the paired samples come from a
process that is still changing, the stationarity precondition stated for
transfer entropy in Appendix~\ref{sec:te} applies to MI as well. Four items form the minimum report. State the estimator family and its key
hyperparameters. Show robustness across $k$ or across training seeds. Give a
bootstrap variance. And say explicitly whether the estimate is a measurement or
a training surrogate.

A near-zero MI value calls for diagnosis. Dismissing it is the mistake, and the
marginal entropies make the diagnosis routine. Since $I(X;Y) \leq
\min\{H(X), H(Y)\}$, a small value can mean either that the two variables are
genuinely close to independent or simply that one of them barely varies: an
agent whose action is almost constant has $H(\mathrm{action}) \approx 0$ and
therefore cannot exhibit appreciable MI with \emph{anything}, whatever it
observes. Reporting the marginal entropies alongside MI separates the two
cases without additional computation, and Section~\ref{subsec:other-measures} shows the
distinction mattering in practice.

\subsection{Transfer Entropy}
\label{sec:te}

\paragraph{Definition and role}
Transfer entropy (TE) from process $Y$ to process $X$ is defined as the
conditional MI $T_{Y \to X} = I(X_{t+1}; Y_t^{(k)} | X_t^{(l)})$, where $Y_t^{(k)}$
and $X_t^{(l)}$ are embedding vectors of lag length $k$ and $l$
respectively~\cite{schreiber2000}. TE measures the directed, time-asymmetric
reduction in uncertainty about $X$'s future given $Y$'s past, above and beyond
what $X$'s own past already provides. It is a model-free measure of directed
predictive dependence. In AI, TE is used to infer directed influence between
components of learned dynamical systems (e.g., between RNN units), to quantify
information routing in modular architectures, and to detect leadership between
time-series streams. Active Information Storage $\mathrm{AIS}(X) =
I(X_{\mathrm{past}}; X_{t+1})$, the self-prediction analogue of TE, is
computed by the same toolboxes~\cite{lizier2014}. In neuroscience, TE
identifies directed information flow between brain regions without assuming
a linear model~\cite{vicente2011,wibral2014,ma2013}.

\paragraph{Estimator choice}
Discrete binning is feasible for low-state-count symbolic data. For continuous
time series, kNN estimators analogous to KSG are the usual nonparametric
choice and are implemented in JIDT~\cite{lizier2014} and
IDTxl~\cite{wollstadt2019}. The critical
hyperparameter is the embedding: uniform embedding uses a fixed lag $l$ and
$k$, while non-uniform embedding selects lags adaptively, reducing redundancy
and improving detection sensitivity~\cite{wollstadt2019}. Where the lag itself
is the quantity of interest, TE has been used to reconstruct interaction delays
directly~\cite{pampu2013}. For spike-train data
(event-based, continuous time), dedicated estimators that avoid binning
entirely are available~\cite{shorten2021}. IDTxl automates embedding selection
via conditional independence testing, which makes it a reasonable default for
multivariate TE analysis when the embedding is not known in advance.

\paragraph{Failure modes and guardrails}
TE is not causality in the interventional sense~\cite{lizier2010,james2016,pearl2009}. An
unobserved common driver can induce positive TE between two processes that
have no direct causal link. Information transfer and causal effect are distinct
quantities that need not agree. They are not coarse and fine versions of one
thing~\cite{lizier2010}. Stationarity violations (trends, seasonality) produce
spurious TE. The minimum reporting protocol has four parts. Give the time lags and the
embedding strategy, with a justification of how they were selected. State the
conditioning set and why it was chosen. Report surrogate significance tests
with effect sizes. Discuss the plausible confounders and how they were
addressed.

The stationarity requirement in particular is not a formality. Standard
estimators pool observations over time, so a violation biases the estimate
directly. In neuroscience this obstacle was severe enough to motivate
ensemble-based estimators that avoid pooling altogether~\cite{wollstadt2014},
and the pitfalls of applying these measures to collectives have been analysed
separately~\cite{sattari2022}. It is tempting to treat the choice of surrogate
construction as the place where such problems get handled.
Section~\ref{sec:experiment} tests that assumption directly, on two of the
standard constructions and on series that share a non-stationary trend. The
practical rule that follows from those measurements is to choose a null model
that respects the non-stationarity, and Section~\ref{sec:experiment} reports
which of four constructions does.

\section{Choosing a Measure and Reporting It}
\label{sec:synthesis}

\subsection{Selecting a Measure}

Table~\ref{tab:master-decision} yields a first-pass
selection. The input is the practitioner's primary objective plus the data
type and dimensionality. The output is a candidate measure, an estimator
family with its key hyperparameters, and the guardrail that goes with it. It
is a starting point to be checked against the failure modes in the measure
subsections. It does not replace reading them.

\subsection{The Four Measures Compared}
\label{subsec:mastertable}

Table~\ref{tab:master-decision} consolidates the four measures across five
dimensions: the measure itself, the decision question it answers, a
representative AI use-case, the recommended estimator(s), and the main
caveat. The estimator column lists measurement tools for analysis. Variational
lower bounds such as MINE~\cite{belghazi2018} and InfoNCE~\cite{oord2018} are
training surrogates unless separately validated on the target distribution,
and should not be silently conflated with
measurement estimators. The caveat column highlights the single most
consequential misuse for each measure.

\begin{table*}[htbp]
  \centering
  \footnotesize
  \caption{The four information-theoretic measures covered in this paper,
    compared across the question each answers, its estimator, and its
    principal failure mode.
    Use-cases and estimators are representative;
    see the corresponding section for a full discussion.
    ``Estimator'' refers to measurement tools;
    neural/variational bounds (MINE, InfoNCE) are training surrogates unless
    separately validated. Toolboxes: JIDT~\protect\cite{lizier2014};
    IDTxl~\protect\cite{wollstadt2019};
    dit~\protect\cite{james2018}.}
  \label{tab:master-decision}
  \renewcommand{\arraystretch}{1.05}
  \setlength{\tabcolsep}{4pt}
  \begin{tabularx}{\linewidth}{@{} p{2.2cm} p{3.6cm} p{3.6cm} p{3.4cm} X @{}}
    \toprule
    \textbf{Measure} &
    \textbf{Question answered} &
    \textbf{AI use-case} &
    \textbf{Estimator} &
    \textbf{Main caveat} \\
    \midrule

    Entropy $H(X)$ &
    How uncertain / diverse is this distribution? &
    Decision trees; max-entropy RL; uncertainty quantification &
    Spacings ($d{=}1$); kNN ($d{\geq}2$); plugin + Miller-Madow (discrete) &
    Bias direction differs between discrete and continuous; report estimator, $k$, and bootstrap CI \\[4pt]

    KL / CE &
    How much does model $q$ differ from truth $p$? Which direction? &
    Cross-entropy loss; VAE regularizer; TRPO/PPO constraint &
    Plugin (discrete); RKHS discriminator (continuous) &
    KL is asymmetric; undefined for disjoint support; distinguish forward vs reverse \\[4pt]

    Mutual Information $I(X;Y)$ &
    How statistically dependent are $X$ and $Y$? (captures non-linear dependence) &
    Feature selection; self-supervised learning; IB tradeoff &
    KSG (low to moderate $d$), GMM-MI (parametric), neural bounds as training
    surrogates only &
    Curse of dimensionality, no single safe cutoff; neural bounds are not
    unbiased estimators; control leakage \\[4pt]

    Transfer Entropy $T_{Y\to X}$ &
    Does $Y$'s past reduce uncertainty about $X$'s future beyond $X$'s own past? &
    Directed influence in RNN / multi-agent; effective connectivity &
    kNN (JIDT, IDTxl) for continuous data, discrete plug-in with Miller-Madow
    for small alphabets; non-uniform embedding recommended &
    Not interventional causality; confounders produce spurious TE; surrogate
    tests required, and a known non-stationary transient must be treated before
    them (Section~\ref{sec:experiment}) \\

    \bottomrule
  \end{tabularx}
\end{table*}

\subsection{Estimators and Software}

Three actively maintained open-source toolboxes cover the measures in
Table~\ref{tab:master-decision}.
\textbf{JIDT}~\cite{lizier2014} (Java, with Python wrapper) implements TE,
MI, and Active Information Storage with KSG and KDE estimators, and is widely
used in neuroscience and dynamical-systems work.
\textbf{IDTxl}~\cite{wollstadt2019} (Python) adds multivariate TE with
automated non-uniform embedding selection and permutation-based significance
testing, which suits multivariate analyses where the embedding must be chosen
from data.
\textbf{dit}~\cite{james2018} (Python) covers discrete IT quantities including
partial information decompositions, useful for symbolic or low-cardinality
state systems.

\paragraph{Reporting minimum}
\label{par:reporting-minimum}
For any IT measure: (R1) specify whether the quantity is a \emph{measurement
estimator} or a \emph{training surrogate}, (R2) report estimator family,
hyperparameters, and software version, (R3) provide bootstrap or subsampling
uncertainty intervals, (R4) include surrogate/permutation significance tests
for TE and MI, (R5) state any preprocessing steps that could affect the
estimate (normalization, discretization, embedding choice), and, for TE
specifically, (R6) treat any known non-stationary transient before testing against
surrogates, by permuting the source within blocks of training time, by excluding it, by
conditioning on the driver, or by estimating across an ensemble of runs, and report which
was done and on what window.
Section~\ref{subsec:checklist} applies these six items to our own experiment.
R6 is the one this paper puts under experimental test, and it is stated as an action. A
disclosure would not have covered it, because what we tested is the doing: a reader can
report a window faithfully and still pool across a transient. The other five are stated practice and are
not validated here.

\section*{Acknowledgment}
In accordance with IEEE Access policy on generative AI, the authors disclose
that in preparing this manuscript they used Claude (Anthropic) for language
editing, grammar, formatting and syntactic revision of text written by the
authors, and Manus (1.6 lite) for workflow automation. Claude was also used,
under the authors' direction and review, to implement the verification
infrastructure released with this paper: the regression guard that recomputes
every reported figure from the stored results, the consistency checks over the
manuscript's numbers and claims, and the version-controlled workflow in which
each change is checked before it is committed. That code is released in full
and its assertions are executable, so any reader can run it against the data
and confirm what it verifies. No section consists of AI-generated text
presented as the authors' own composition, so no section carries an AI-system
citation. The experimental design, the analysis, the interpretation of results
and all scientific conclusions are the authors', who reviewed and edited every
output and take full responsibility for the content of this publication.

\section*{Data Availability}
All data underlying the experiment in Section~\ref{sec:experiment} were
generated by the authors and are available, together with the complete
analysis code and the seed sweep, as supplementary material. They are archived at
\url{https://doi.org/10.5281/zenodo.22658530} and the code is also maintained at
\url{https://github.com/dentros/te-nonstationarity}. The multi-agent
environment and agent implementations used to generate that data are supplied
by the open-source, GPLv3-licensed GameBrains platform~\cite{gamebrains2026},
publicly available at \url{https://github.com/dentros/gamebrains}. All other
tools and reference implementations cited are publicly available and listed in
the bibliography.

\section*{Conflict of Interest}
The authors declare no conflict of interest. The work received no external
funding, and no funder had any role in the design of the experiment, the
analysis or the interpretation of the results, or in the decision to publish.

\bibliographystyle{IEEEtran}
\bibliography{references}

\begin{IEEEbiographynophoto}{Nikolaos Al. Papadopoulos}
received the B.Sc. degree in Applied Informatics, Technology Management
direction, from the University of Macedonia, Thessaloniki, Greece, and the
M.Sc. degree in Cognitive Systems and Interactive Media from Universitat
Pompeu Fabra, Barcelona, Spain. Currently pursuing the Ph.D. degree with the
Department of Applied Informatics, University of Macedonia. Research interests
include information-theoretic analysis of learning systems, multi-agent
reinforcement learning, computational models of decision-making agents, and
reproducible experimental methodology for artificial intelligence.
\end{IEEEbiographynophoto}

\begin{IEEEbiographynophoto}{Konstantinos E. Psannis}
received the degree in physics from Aristotle University of Thessaloniki,
Greece, and the Ph.D. degree from the Department of Electronic and Computer
Engineering, Brunel University, London, U.K., under a British Chevening
Scholarship. Currently a Full Professor of Communication Systems and Networking with the
Department of Applied Informatics, University of Macedonia, Thessaloniki,
Greece, and a Visiting Consultant Professor with the Nagoya Institute of
Technology, Japan. Also Director of the Mobility2net JP-EU Laboratory and a
member of the EU-Japan Centre for Industrial Cooperation. Research spans
digital media communications and transport over heterogeneous networks,
6G-enabled big data, AI-IoT and digital twins. Author of more than 100 journal
articles, 150 conference papers and 30 book chapters, and listed among the
Stanford University top 2\% most-cited researchers.
\end{IEEEbiographynophoto}

\EOD

\end{document}